\pdfoutput=1

\documentclass[11pt]{article}

\usepackage[final]{acl}
\usepackage[most]{tcolorbox}
\tcbuselibrary{listingsutf8} %
\usepackage{CJKutf8}         %
\usepackage{times}
\usepackage{latexsym}
\usepackage{amsmath}
\usepackage{amssymb}
\usepackage{bbm}
\usepackage[T1]{fontenc}

\usepackage[utf8]{inputenc}
\usepackage{svg}

\usepackage{microtype}

\usepackage{inconsolata}
\usepackage{booktabs}
\usepackage{multirow}
\usepackage{tcolorbox}
\usepackage{graphicx}
\usepackage{stmaryrd}
\usepackage{amsfonts}
\usepackage{amssymb}
\usepackage{amsmath}
\usepackage{bbm}
\usepackage{xspace}
\usepackage{todonotes}
\usepackage[linesnumbered,ruled,vlined]{algorithm2e}
\usepackage{tabularx,booktabs,multirow}  
\usepackage{subcaption}
\usepackage{enumerate}
\usepackage{siunitx}
\usepackage{amsthm}

\usepackage{amsmath,amsfonts,bm}

\def\eqref#1{equation~\ref{#1}}

\def\1{\bm{1}}

\def\ve{{\bm{e}}}

\def\vs{{\bm{s}}}

\def\mM{{\bm{M}}}

\DeclareMathAlphabet{\mathsfit}{\encodingdefault}{\sfdefault}{m}{sl}
\SetMathAlphabet{\mathsfit}{bold}{\encodingdefault}{\sfdefault}{bx}{n}

\theoremstyle{definition}

\theoremstyle{remark}

\title{Shared Path: Unraveling Memorization in Multilingual LLMs through Language Similarities} 

\author{Xiaoyu Luo\textsuperscript{1,2},\space
  Yiyi Chen\textsuperscript{1},\space
  Johannes Bjerva\textsuperscript{1},\space
  Qiongxiu Li\textsuperscript{2}\thanks{\ \ Corresponding author.}\\
  \textsuperscript{1}Department of Computer Science, \textsuperscript{2}Department of Electronic Systems\\ 
  Aalborg University, Copenhagen, Denmark\\
\texttt{\{xilu,yiyic,jbjerva\}@cs.aau.dk,qili@es.aau.dk}
  }

\begin{document}
\maketitle
\begin{abstract}
We present the first comprehensive study of \textit{Memorization} in Multilingual Large Language Models (MLLMs), analyzing 95 languages using models across diverse model scales, architectures, and memorization definitions. 
As MLLMs are increasingly deployed, understanding their memorization behavior has become critical. 
Yet prior work has focused primarily on monolingual models, leaving multilingual memorization underexplored, despite the inherently long-tailed nature of training corpora. 
We find that the prevailing assumption, that memorization is \textit{highly correlated} with training data availability, fails to fully explain memorization patterns in MLLMs.
We hypothesize that the conventional focus on monolingual settings, effectively treating languages in isolation, may obscure the true patterns of memorization.
To address this, we propose a novel graph-based correlation metric that incorporates language similarity to analyze cross-lingual memorization. 
Our analysis reveals that among similar languages, those with fewer training tokens tend to exhibit higher memorization, a trend that only emerges when cross-lingual relationships are explicitly modeled. 
These findings underscore the importance of a \textit{language-aware} perspective in evaluating and mitigating memorization vulnerabilities in MLLMs. 
This also constitutes empirical evidence that \textit{language similarity} both explains \textit{Memorization} in MLLMs and underpins \textit{Cross-lingual Transferability}, with broad implications for multilingual NLP~\footnote{We release our code at: \url{https://github.com/xiaoyuluoit97/MLLM_memorization}.}.

\end{abstract}

\section{Introduction}
Large Language Models (LLMs) demonstrate increasingly strong capabilities in processing and understanding multiple languages \cite{conneau-etal-2020-unsupervised}, resulting in advancements across a wide range of natural language processing (NLP) tasks \cite{choi2021analyzingzeroshotcrosslingualtransfer,pikuliak2021cross}. 
MLLMs, in particular, empower global users to interact in their native languages, offering wide-reaching benefits in accessibility and productivity.  
\begin{figure}[!t]
    \centering
    \includegraphics[width=0.95\linewidth]{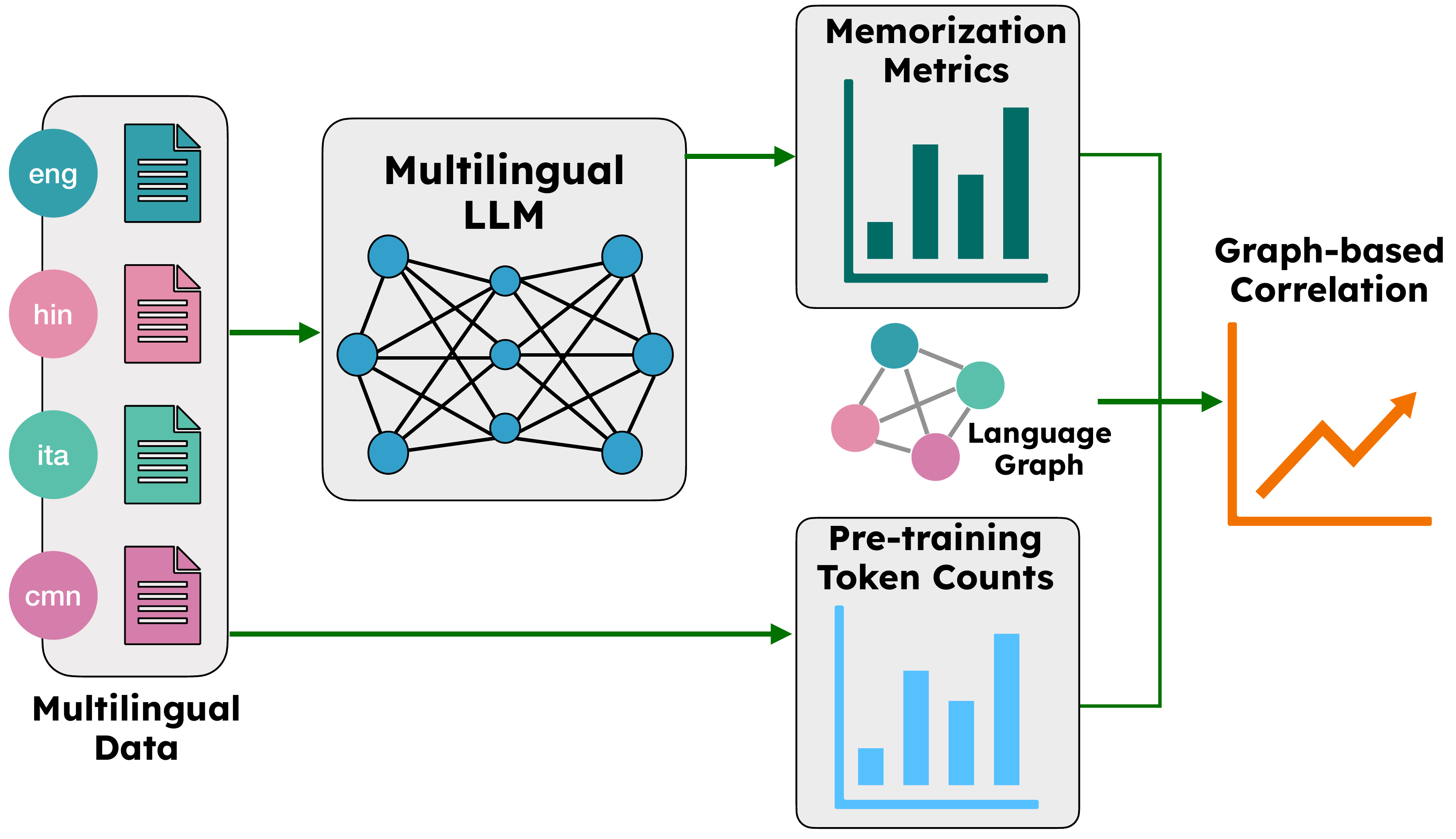}   
    \caption{Overview of our Framework for Analyzing Memorization in MLLMs using Language Similarity Graph-based Correlation Analysis.}
    \vspace{-5pt}
    \label{fig:figure1}
\end{figure}

However, LLMs are also known to \textit{memorize} portions of their training data~\cite{carlini2021extracting}, raising serious concerns such as the leakage of copyrighted content~\cite{chang2023speak} and personal information~\cite{staabbeyond}. While memorization in monolingual LLMs has been widely studied, how it manifests in multilingual models remains \textit{underexplored}.

Prior work predominantly attributes memorization to data volume, positing that frequent tokens or duplicated content are disproportionately memorized ~\cite{carlini2022quantifying}. This echoes findings from computer vision, where long-tail examples are disproportionately memorized~ \citep{feldman2020neural,jiang2020exploring,garg2023memorization}, resulting in increased privacy and fairness risks~\cite{li2024privacy, gao2023not, tramer2022truth}. 
However, MLLMs introduce a unique complexity: languages are not processed independently but in a joint space, often sharing lexical, morphological, and syntactic features.
While prior memorization research has largely focused on monolingual models and settings \citet{carlini2021extracting, carlini2022quantifying}, without explicitly examining the role of cross-lingual similarity, our work explores how such relationships may shape memorization dynamics.
For instance, typologically similar languages like Turkish and Azerbaijani may interact during training in ways that affect their memorization patterns. Moreover, low-resource languages naturally occupy the long tail of the data distribution, introducing complex dynamics that are poorly understood. 
Together, these challenges raise important questions that motivate our investigation. For example, to what extent does memorization in MLLMs correspond to training data volume, as suggested by long-tail distribution assumptions? How might cross-lingual relationships influence memorization behavior across languages? And can memorization in one language lead to unintended leakage in another, particularly among similar languages?

To answer these questions, we conduct the first large-scale study of memorization in MLLMs,  uncovering critical limitations of existing research and offering a novel language-aware perspective (see Fig.~\ref{fig:figure1} for an overview of our framework).  
Our key contributions are:
\begin{itemize}
\item \textbf{Revisiting the Long-Tail Assumption}: We show that memorization in multilingual settings cannot be fully explained by training data volume or token frequency. In many cases, low-resource languages exhibit lower memorization rates than high-resource counterparts.
\item  \textbf{Language Similarity-Aware Correlation Metric:} We introduce a novel graph-based correlation metric that incorporates typological and statistical similarities between languages, enabling structured analysis of cross-lingual memorization dynamics.
\item \textbf{Cross-Lingual Memorization Insights:} Using our metric, we find that languages with high similarity exhibit interconnected memorization behaviors, affording fundamental grounding for cross-lingual transferability. 
\item \textbf{Comprehensive and Robust Evaluation:} We assess memorization using both generation-based and likelihood-based metrics, and validate our findings across over 95 languages, multiple LLM architectures (encoder-only and decoder-based) of varying scales, demonstrating consistent and generalizable trends.
\end{itemize}

\section{Related work}
\subsection{Memorization in LLM}
Memorization in deep neural networks has long been recognized as a critical issue, with implications for privacy, fairness, and generalization \cite{feldman2020neural,garg2023memorization,chang2021privacy,li2025trustworthy}.
These concerns have been empirically confirmed in LLMs. \citet{carlini2019secret} first show that generative models can inadvertently memorize and reproduce rare, sensitive training data. \citet{carlini2021extracting} further demonstrate that large models like GPT-2 can regurgitate unique sequences even if they appear only once in the training corpus. \citet{carlini2022quantifying} systematically quantify memorization patterns across model scales and architectures, while \citet{kim2023propile} focus on personally identifiable information (PII) memorized by LLMs, proposing ProPILE to assess leakage from the perspective of data subjects.

Recent work has formalized memorization risk, particularly distinguishing between \textit{discoverable} and \textit{extractable memorization}~\cite{carlini2021extracting, nasr2023scalable}. The latter refers to information that an adversary can extract without direct access to the training set, posing realistic threats to deployed models.
Studies have shown that LLMs, including GPT, T5, and others, can leak hundreds to millions of training sequences depending on model sizes, data duplication, and prompt strategies~\cite{nasr2023scalable, carlini2022quantifying}.
Beyond quantifying leakage, several studies have advanced the understanding of memorization mechanisms and measurement approaches. \citet{chen2024multi} analyze how model and context size affect transitions between unmemorized and memorized outputs. \citet{liu2024forgetting} propose the forgetting curve, a corpus-agnostic method to reliably measure memorization capability across architectures. \citet{li2024rome} introduce ROME, revealing how token length and prediction confidence relate to memorization without relying on training data access. \citet{ haviv2022understanding} demonstrate that recall of memorized sequences follows a two-stage process of early promotion and later confidence amplification in transformer models. \citet{stoehr2024localizing} localize memorization to specific low-layer attention heads and high-gradient parameters, showing such content is harder to unlearn.
While such risks have been studied in monolingual settings, memorization behavior in multilingual LLMs remains underexplored, with the exception of~\citet{cavalin-etal-2024-fixing}, especially for low-resource languages occupying the long tail of the training distribution.

\subsection{Cross-lingual Transferability \& Language Similarity}
Cross-lingual transfer entails the representation of texts in multiple natural languages in a shared multilingual space. 
The paradigm of representations for cross-lingual transfer has shifted from word embeddings~\citep{mikolov2013efficient,ammar2016many,vulic-etal-2019-really} to contextual embeddings~\citep{conneau2019unsupervised, devlin-etal-2019-bert, raffel2020exploring}.
Previous work investigating cross-lingual transferability mainly leverages downstream task performance to measure the transfer from a source language or languages to target languages through selective fine-tuning~\citep{choenni2023languages} or using zero-shot or few-shot transfer with pre-trained MLLMs~\citep{Lauscher2020FromZT,adelani-etal-2022-masakhaner,de-vries-etal-2022-make, blaschke2025analyzing}.
Language similarity based on linguistic data has been heavily referred to in cross-lingual transferability studies~\citep{wichmann2011phonological,littell2017uriel}, not without faulty representations~\citep{Toossi2024ARS, khan-etal-2025-uriel}.
Moreover, the findings on leveraging language similarity for improving downstream cross-lingual transfer remain mixed and sometimes contradictory~\citep{philippy2023towards}. 
Recently, different language similarity measures have been deployed to enhance cross-lingual transfer performance under different NLP tasks~\citep{blaschke2025analyzing} and analyze MLLM language distribution patterns~\citep{chen-etal-2025-large}.
We share the perspective that \textit{language similarity is not a static concept}, and different measures can be pertinent to different scenarios.

Prior research in MLLM embedding spaces has shown that sentence embeddings are composed of a \textit{language-specific} and \textit{language-agnostic} components ~\citep{pires-etal-2019-multilingual, libovicky-etal-2020-language, xie2024discovering}, which have been leveraged to improve downstream performance~\citep{tiyajamorn-etal-2021-language} and investigate language relations in MLLMs~\citep{choenni2022investigating}. 
In addition,~\citet{lin-etal-2024-mplm} shows that language similarity extracted from pretrained MLLMs with parallel sentences exhibits moderately high correlations with linguistic similarity measures, further motivating our language-aware memorization analysis.
Notably, \citet{zhao2024word} demonstrate that even within closely related languages, structural factors such as word order can yield divergent outcomes in knowledge induction, underscoring that language similarity is multifaceted and context-dependent.
In this paper, we extract language-specific embeddings from each MLLM as language representations to compute language similarity (cf. Section~\ref{measuring_language_simialrity}).



\section{Language Model Memorization~\label{sec:definition_memorization}}

We define \textit{Memorization} in the context of LLMs and examine its key formulations from different perspectives.
Given an LM $f$ and a string $x$ from its training data, we split $x$ into a prefix $p$ and a suffix $s$, so that \(x = p || s\).
Let the prefix $p$ consist of $n$ tokens, noted as $p=(p_1, \dots, p_n)$;
and let the suffix $s$ consist of $m$ tokens, noted as $s=(s_1, \cdots, s_m) $.

\subsection{Measuring MLLM Memorization}
\paragraph{Exact Memorization} Following the definition of extractable memorization by~\citet{carlini2022quantifying}, whether a language model can reproduce a training sequence when prompted with part of it using greedy decoding, 
we define \textit{Exact Memorization Ratio} as \(\frac{n}{n+m}\) to measure the fraction of the sequence required for exact reconstruction.
Given a set of samples, 
we define the \textit{Exact Memorization Rate (EM)} as the fraction of samples where the model, when prompted with the prefix, reproduces the suffix exactly:
\vspace{-5pt}
\[
\text{EM} = \frac{1}{N} \sum^{N}_{i=1} \mathbbm{1}(\hat{s_i}=s_i),
\vspace{-5pt}
\]
where $N$ is the total number of samples, $s_i$ is the true suffix of the $i$th sample, $\hat{s}_i$ is the output given the prefix and $\mathbbm{1}(\cdot)$ is the indicator function.

\paragraph{Relaxed Memorization}
As \textit{Exact Memorization} is a stringent criterion, we additionally define a relaxed version of memorization that evaluates the predicted suffix against the ground truth suffix using approximate string matching metrics rather than exact match.
We use BLEU~\citep{papineni2002bleu} and Rouge-L~\citep{lin2004rouge} as our \textit{Relaxed Memorization Scores (RM)}, serving as continuous indicators of memorization.

\paragraph{Reconstruct Likelihood Memorization}
Complementary to previous generation-based memorization metrics, we adopt reconstruct likelihood from~\citet{kim2023propile} to define a probability-based metric \textit{Reconstruct Likelihood Memorization}, noted as \textit{PM}.
which quantifies memorization by the likelihood the model assigns to a known sequence under its learned distribution, i.e., its internal probability of reconstructing the suffix given its prefix.
Our goal is to evaluate how likely the model finds the suffix $s$ when conditioned on the prefix $p$. 
We define the log-likelihood of $s$ given $p$ as:
\vspace{-5pt}
\[
\log \Pr(s \mid p) = \sum_{r=1}^{m} \log p(s_r \mid p, s_{<r}),
\vspace{-5pt}
\]
where $s_{<r}$ denotes the preceding $r-1$ tokens of the suffix.

\subsection{Memorization for Encoder-Decoders~\label{encoder-decoder-setup}}

The definitions above primarily assume a decoder-only architecture of LLMs where predictions are made in a left-to-right autoregressive manner. 
In contrast, encoder-decoder models such as T5 are trained with a span-denoising objective~\citep{raffel2020exploring}.
Following~\citet{carlini2022quantifying}, we randomly mask a set of non-contiguous token spans from a sampled data sequence.
To evaluate \textit{Exact Memorization}, the model reconstructs these missing spans given surrounding context, and we consider a string to be memorized if the generated output exactly matches the masked content.
To evaluate \textit{Reconstruct Likelihood Memorization}, we follow the span corruption setup and treat the masked spans as targets. 
We then compute the sum of log-probabilities assigned to these tokens, conditioned on the visible parts of the sequence.

T5's span corruption objective typically mask very short spans (about three tokens on average under default settings~\citep{raffel2020exploring}), so token-level similarity becomes uninformative, hence we do not assess the relaxed memorization for T5-based encoder-decoder models.

\section{Methodology}
\begin{figure}[t!]
    \centering
    \includegraphics[width=\linewidth]{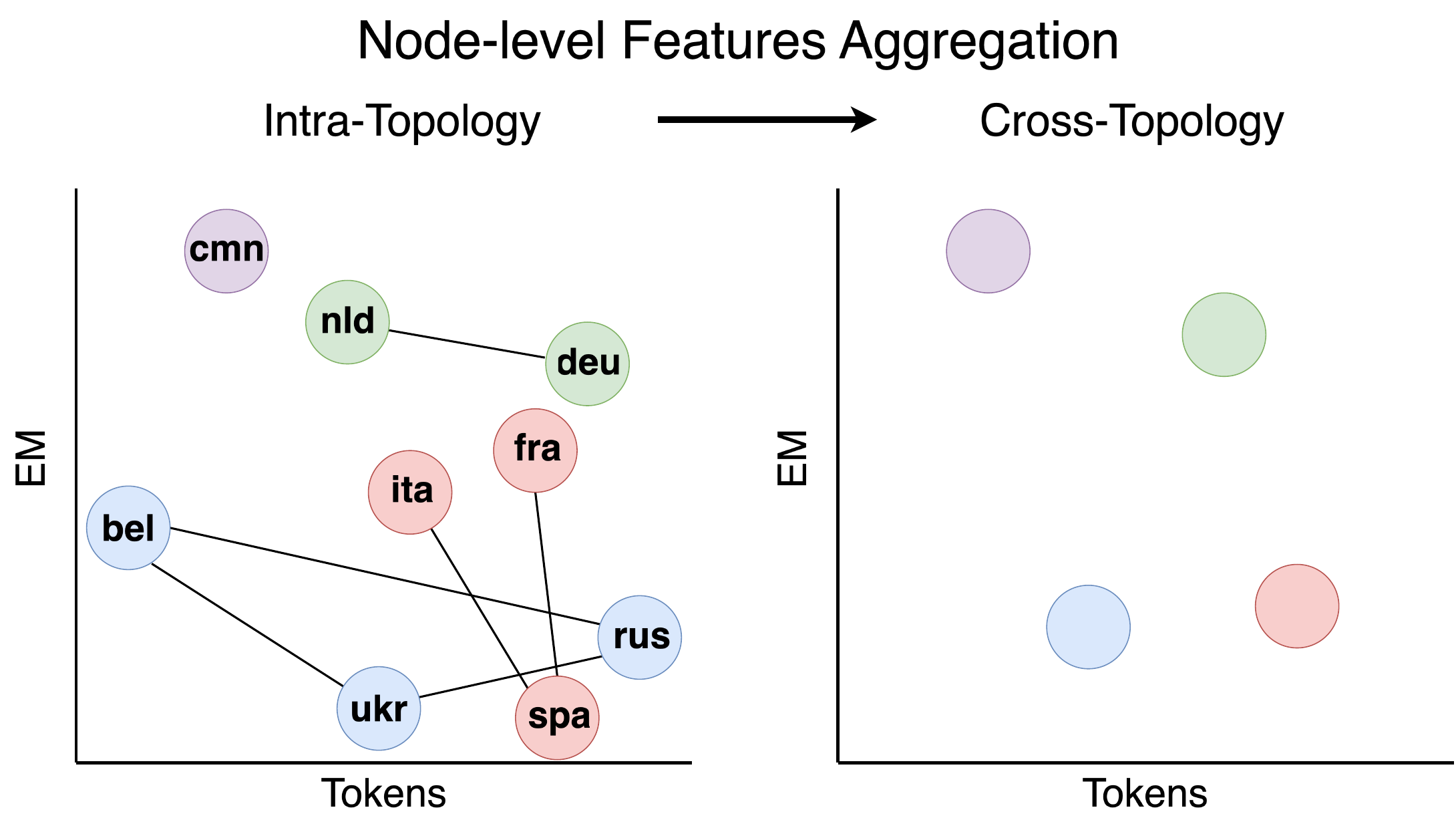} 
    \caption{Example graphs considering Intra-Topology and Cross-Topology.}
    \vspace{-5pt}
    \label{fig:intra_cross_corr}
\end{figure}

Previous work on LLM \textit{Memorization} has mainly focused on data duplication and frequency in monolingual settings, with limited analysis across languages. Although correlation metrics such as Pearson can quantify global trends (e.g., measuring how  token counts and memorization rates linearly co-vary), they overlook the structured dependencies among languages.
Our analysis (Fig.~\ref{fig:intra_cross_corr}) shows that languages with similar frequency distributions can exhibit divergent memorization patterns, underscoring the importance of language-aware evaluation.

\subsection{Measuring Language Similarity~\label{measuring_language_simialrity}}
We leverage \textit{language-specific} subspace in multilingual embedding space to measure language similarities. 
Let $L$ be a set of languages.
To extract language representations from MLLMs, we use a parallel dataset $D$, in our case Flores+~\citep{nllb-24}, which is entirely separate from the model’s training data and contains 2,000 examples per language. 
Suppose we have $m$ sentences for each language $l\in L$ in $D$, we first extract the mean embedding $\mu_l= \frac{1}{m}\sum^{m}_{i=1} \ve_{l}^{i}$ for each hidden layer $h$, where $\ve_{l}^{i}\in \mathbb{R}^{d} $ is a sentence embedding.
We then form a matrix $M \in \mathbb{R}^{d \times |L|}$ by concatenating $\mu_l$ across all languages.
We extract the language-specific subspace $M_s$ using Algorithm~\ref{algo:lsar}~\citep{xie2024discovering} (see Appendix~\ref {sec:langsubspace} for details), then project each language embedding into this subspace $\vs_l = M_s M_s^{T} \ve_l$.
For each hidden layer $h$ in a MLLM, we measure the pair-wise language similarity for a language pair $\{l_1, l_2\}$, where \def\ve{{\bm{e}}}
$l_1, l_2\in L $ using cosine similarity between language-specific embeddings:
\vspace{-5pt}
\[
\cos{(\vs_{l_{1}},\vs_{l_{2}})}= \frac{\vs_{l_{1}} \cdot \vs_{l_{2}}}{||\vs_{l_{1}}||\cdot  ||\vs_{l_{2}}||}.
\vspace{-5pt}
\]
Empirically, we find that the language similarity drawn from the final layer embeddings of MLLMs shows a stronger correlation with linguistically grounded similarity measures overall (cf. Appendix~\ref{app:layer-correlation}).

\subsection{Graph-based Correlation Analysis}
We introduce our topology-based framework, which captures cross-lingual dependencies by modeling signal propagation over a language similarity graph. 
It rests on two empirical observations:  
(1) Memorization patterns tend to propagate across related languages, and  
(2) Standard correlation metrics fail to capture these structured transfer effects.

\paragraph{Graph Construction via Language Similarity}
We represent the language space as an undirected graph $\mathcal{G} = (\mathcal{V}, \mathcal{E})$, where each node corresponds to a language, and edges encode pairwise language similarity. 
Let $n$ be the number of languages and $A \in \mathbb{R}^{n \times n}$ the adjacency matrix, where $A_{ij}$ represents the similarity between languages $i$ and $j$. To sparsify the graph and remove self-loops, we apply thresholding with $\theta$:
\vspace{-5pt}
\begin{align}\label{eq.theta}
    A_{ij} =
\begin{cases}
1, & \text{if } \text{sim}(i, j) \geq \theta \\
0, & \text{otherwise}
\vspace{-5pt}
\end{cases}.
\end{align} 

We then construct the unnormalized graph Laplacian matrix $L = D - A$, where $D_{ii} = \sum_j A_{ij}$ is the degree matrix.

\paragraph{Information analysis over the Graph}
To understand how language-level signals behave over this graph structure, we begin with the concept of \textit{graph smoothness}, which quantifies how much a signal varies across adjacent nodes. For a scalar-valued signal $\mathbf{x} \in \mathbb{R}^n$ defined over graph, the smoothness is defined as~\cite{zhou2004regularization}:
\vspace{-5pt}
\[
\mathbf{x}^\top L \mathbf{x} = \sum_{(i,j) \in \mathcal{E}} A_{ij}(x_i - x_j)^2.
\vspace{-5pt}
\]
Smaller values indicate that the signal $\mathbf{x}$ changes slowly over similar nodes, i.e., it is \textit{smooth} with respect to the graph topology.

To compare how two signals (e.g., memorization scores and the number of tokens) vary together across languages, we define the \textit{graph cross-smoothness}:
\vspace{-5pt}
\[
\mathbf{x}^\top L \mathbf{y} = \sum_{(i,j) \in \mathcal{E}} A_{ij}(x_i - x_j)(y_i - y_j),
\vspace{-5pt}
\]
where $\mathbf{y} \in \mathbb{R}^n$ refers to a scalar-valued signal different from $\mathbf{x}$.
This measures whether the two signals increase and decrease in tandem over topologically similar languages.

\paragraph{Graph-based Correlation Coefficient}
Based on the above definitions, we define the proposed \textit{Graph-based Correlation Coefficient} between signals $\mathbf{m}$ (e.g., memorization scores) and $\mathbf{t}$ (e.g., token counts) as:
\vspace{-5pt}
\[
\rho_G(\mathbf{m}, \mathbf{t}) = \frac{ \mathbf{m}^\top L \mathbf{t} }{ \sqrt{ (\mathbf{m}^\top L \mathbf{m}) (\mathbf{t}^\top L \mathbf{t}) } }
\vspace{-5pt}
\]
Note that the defined coefficient is bounded by the Cauchy-Schwarz inequality:
\vspace{-3pt}
\[
|\mathbf{m}^\top L \mathbf{t}| \leq \sqrt{ (\mathbf{m}^\top L \mathbf{m}) (\mathbf{t}^\top L \mathbf{t}) } \quad 
\vspace{-3pt}
\]
Hence, $ \rho_G(\mathbf{m}, \mathbf{t}) \in [-1, 1]$ and it captures the structural alignment between the two signals over the graph. A value close to 1 implies that memorization and token frequency change similarly across related languages, while values near $-1$ implies inverse alignment.

$\rho_G$ accounts for the topological structure of language space, enabling us to uncover subtle, structure-respecting relationships in MLLM memorization, which would otherwise be missed by flat, language-agnostic analyses such as Pearson correlation (cf. Table~\ref{tab:model-results-exact-mgpt101} for details).

\subsection{Intra-Topology \& Cross-Topology Analysis}
To further interpret the structure of memorization alignment, we partition the graph into subgraphs by thresholding edge weights. Each subgraph represents a cluster of similar languages; disconnected components reflect cross-topological groups.
To enable meaningful comparison across different language topology clusters, we aggregate node-level features into a single representative vector per subgraph. 
This aggregation is performed within each subgraph, it is weighted by language prominence (node degrees) and normalized by global edge weights to preserve topological information. Specifically, for a subgraph $\mathcal G' = (\mathcal V', \mathcal E')$, where each node $i \in \mathcal V'$ has features $t_i$ (tokens) and $m_i$ (memorization), we define the subgraph-level representations as:
\vspace{-5pt}
\[
\begin{aligned}
\bar{t} &= \sum_{i \in \mathcal V'} \left( \frac{n_i}{\sum_{j \in \mathcal V'}n_j} \cdot t_i \right)
\vspace{-5pt}
\end{aligned}
\]
where $n_i = |\{ j \mid (i,j) \in \mathcal E' \}|$ is the degree of node $i$. The aggregated memorization $\bar{m}$ is computed similarly.

We refer \textbf{intra-topo} as the set of language nodes connected by edges in the graph, while \textbf{cross-topo} refers to language groups that remain disconnected.
The resulting subgraph-level representations enable cross-topology correlation analysis via Pearson correlation.
This approach remains faithful to the internal structure of each language cluster, while capturing the relationship between memorization and training tokens across topologically dissimilar clusters. 
It complements our topology-aware metric $\rho_G$ by offering a cluster-level, interpretable view of memorization–complexity alignment.

\section{Experimental Setup}
\subsection{Model Selection \& Corpus Details}
Studying memorization in MLLMs requires i) publicly available models with ii) fully disclosed pre-training data and iii) broad language coverage. 
For fair cross-architecture comparisons, we also align models by their training corpora and tokenizers whenever feasible.
We use the \textsc{mT5} encoder-decoder family~\cite{xue2020mt5}, trained on \textsc{mC4}~\cite{raffel2020exploring} covering 100+ languages, and the \textsc{mGPT} decoder-only series for architectural comparison.
Specifically, \textsc{mGPT-101} shares the tokenizer and mC4 training data with \textsc{mT5-base}.
Additionally, we select \textsc{mGPT-1.3B} and \textsc{mGPT-13B} to assess scale effects, which are trained on more balanced and filtered \textsc{mC4} (cf. Table~\ref{tab:model_selection} for details).

As shown in Fig.~\ref{fig:corpus_distribution_1} and~\ref{fig:corpus_distribution_2} in Appendix~\ref{app:corpus_distribution}, the data distribution of \textsc{mC4} across languages exhibits a clear \textit{long-tailed} pattern. 
A small number of high-resource languages (such as English, Russian, and Spanish) dominate the corpus in terms of token count, while the vast majority of other languages are represented with significantly fewer tokens. This long-tailed distribution serves as an important factor in analysing how memorization behaviors vary across languages in MLLMs.



\subsection{Prompt sampling}
\textsc{mC4} contains a substantial amount of noisy and duplicated content.
For pre-processing, we sample text passages with more than 600 characters, and filter the content containing ``http://'', garbled tokens, repeated strings, and long sequences of meaningless digits. 
To ensure accurate language representation, we use CLD3 \cite{cld3} for language identification. 
Specifically, we retain only those samples where both the predicted language confidence and the proportion of the target language exceed 90\%.

Duplicated content can disproportionately impact memorization,
where sequences that appear more frequently in the training set are more likely to be memorized, following a near log-linear trend~\cite{lee2021deduplicating}. 
To control repetition for minimizing potential bias and ensure a more balanced representation across the dataset, we randomly sample 50,000 filtered examples per language with a 5 million shuffle buffer, following the sample size in~\citet{carlini2022quantifying}. A handful of low-resource languages with insufficient examples are marked with an asterisk and boldface in Fig.~\ref{fig:corpus_distribution_1};~\ref{fig:corpus_distribution_2}.

\section{Analysis \& Results}
We investigate \textit{Memorization} in MLLMs across multiple dimensions: languages, model architectures, prompt length and model scale.
In each dimension, we measure the Memorization Rates (cf. Section~\ref{sec:definition_memorization}) and correlate with training data (in token counts) in languages, using both the Pearson correlation ($r$) and Graph-based Correlation ($\rho_{G}$) metrics.
\begin{figure}[t!]
    \centering
    \includegraphics[width=0.60\linewidth]{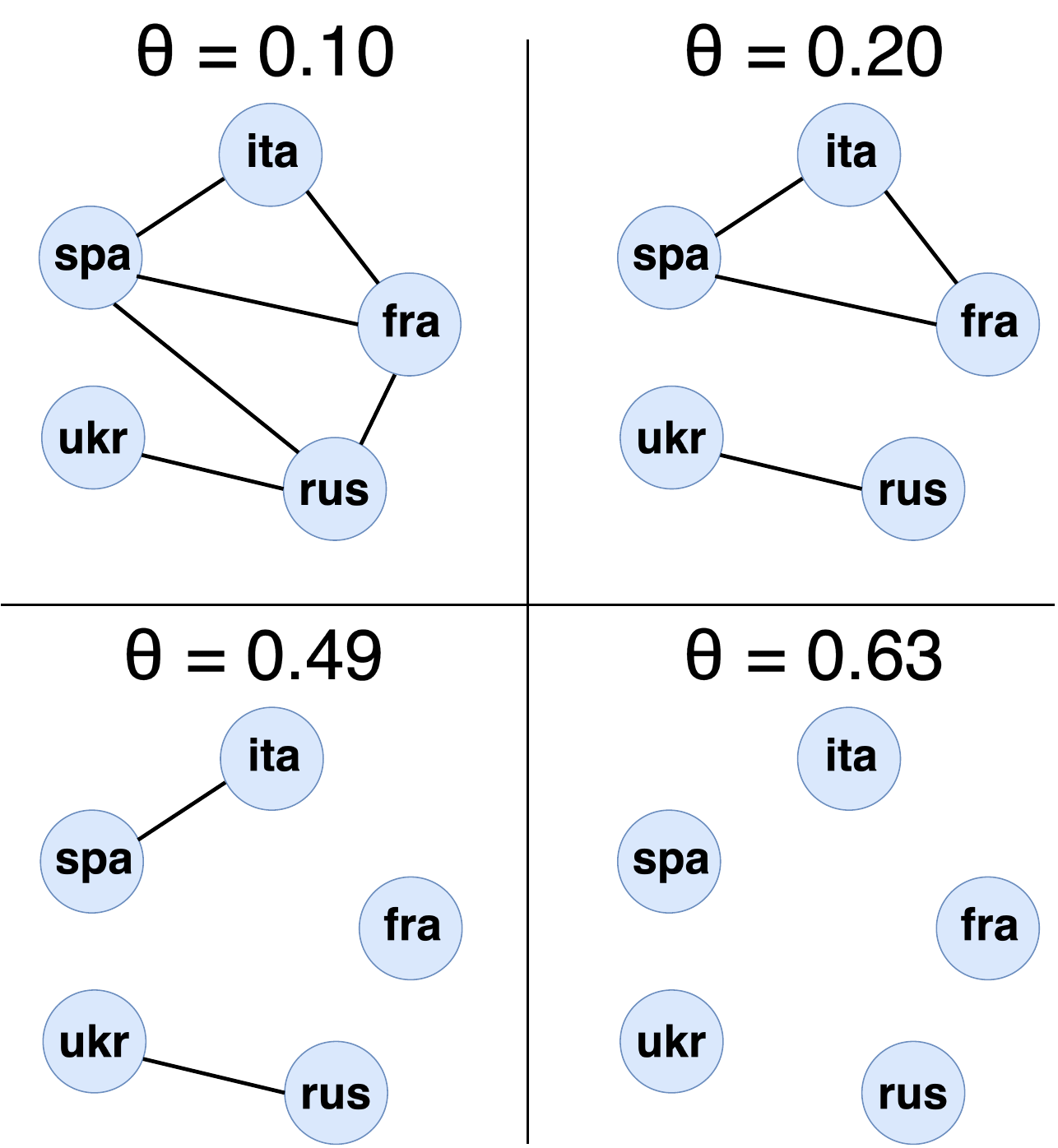}
    \caption{Graph Construction at Different Thresholds $\theta$.}
    \vspace{-5pt}
    \label{fig:threshold_change}
\end{figure}

\subsection{Constructing Language Graphs~\label{sec:graph_construction}}
To use our graph-based correlation to analyze memorization in MLLMs, we construct language similarity-based graphs, at varying thresholds $\theta$ based on ~\eqref{eq.theta},
which specifies the minimum similarity required for two languages to be considered meaningfully related. 
Thus, \(\theta\) directly controls the sparsity of the resulting language graph.
Fig.~\ref{fig:threshold_change} illustrates this effect using a subset of \textsc{mGPT-101} pre-training languages, showing how edge density and connectivity increase as \(\theta\) increases. 
As expected, higher language similarity thresholds $\theta$, the fewer connected graphs. 
By varying \(\theta\), we adjust the granularity of the language similarity topology, enabling analysis under different levels of relational strictness.

\subsection{Data Availability in Memorization~\label{sec:result_llm_mem}}
We evaluate the relationship between per-language memorization rates and the token counts in training data in a MLLM, for example, \textsc{mGPT-101}. 
\begin{table}[t!]
\centering
\small
\begin{tabular}{c|cc}
\toprule
\textbf{Mem. Metric} & \textbf{$r$} & \textbf{$\rho_{G}$} \\
\midrule
EM   & -0.13 & -0.24  \\
PM & -0.36 & -0.56  \\
RM (BLEU)    & -0.23 & -0.36   \\
RM (Rouge-L)  & -0.06 & -0.30   \\
\bottomrule
\end{tabular}
\caption{Correlations between Memorization Rates and Training Data in Token Counts of \textsc{mGPT-101}. The $\rho_G$ with graph-based metric threshold $\theta$ = 0.41.  \textbf{Takeaway:} the proposed $\rho_G$ accentuates the correlation.
}
\vspace{-5pt}
\label{tab:model-results-exact-mgpt101}
\end{table}
As shown in Table~\ref{tab:model-results-exact-mgpt101},
our graph-based metric $\rho_{G}$, by incorporating language similarity, largely \textit{accentuates} the negative correlation between language-wise memorization and token count, in comparison to the Pearson correlation coefficient.
This negative trend suggests that, among similar languages, those with fewer training tokens tend to exhibit higher memorization, 
which further corroborates our hypothesis that memorization in MLLMs cannot be explained by training data volume alone. 



\subsection{Cross-lingual Transferability vs. Memorization~\label{sec:cross-topo}}
\begin{figure}[h]
    \centering
    \includegraphics[width=\linewidth]{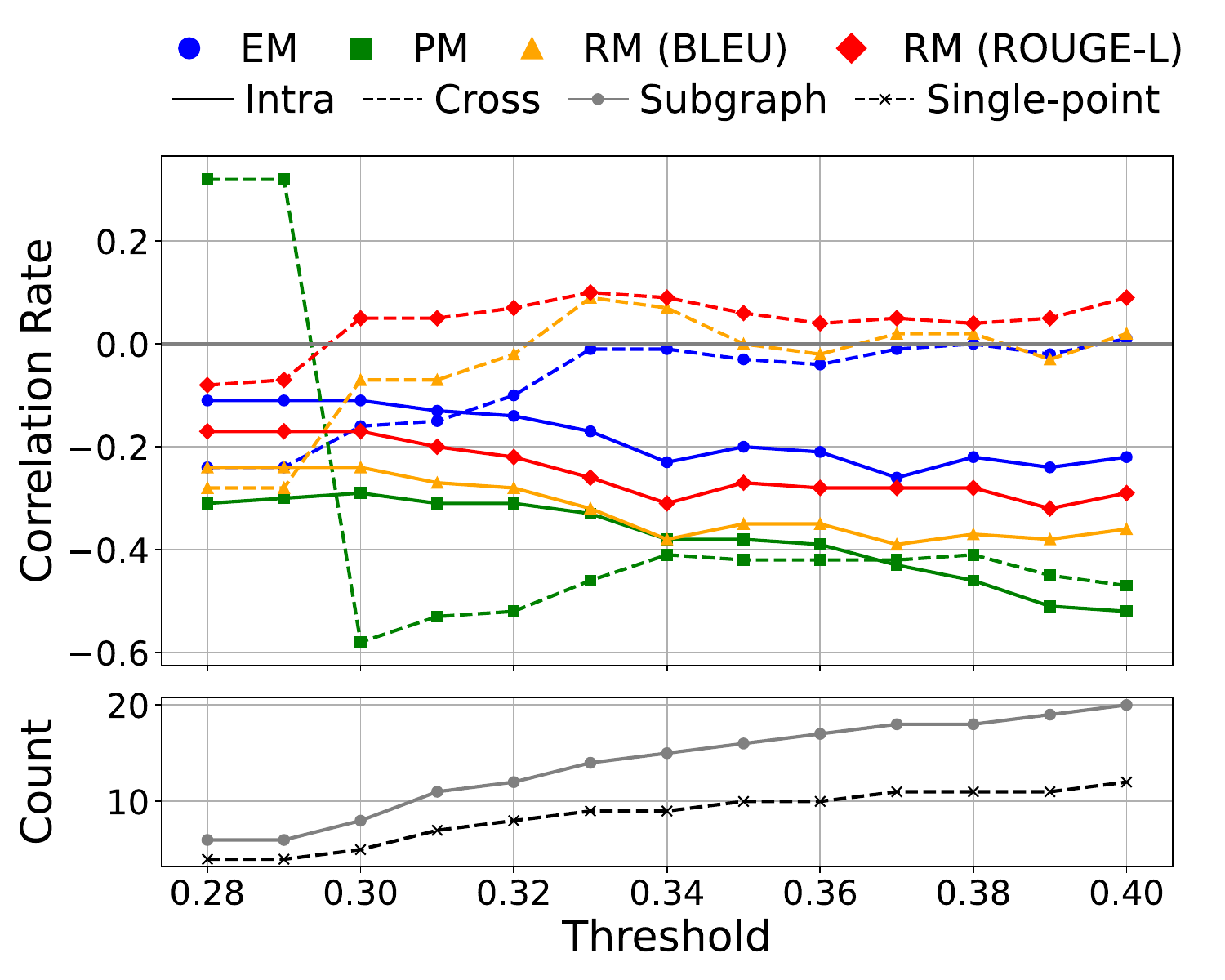}
    \caption{Intra-Topology and Cross-Topology Correlation Coefficients ($\rho_G$) across varying thresholds~$\theta$. \textbf{Top}: Memorization Rates across Thresholds. \textbf{Bottom}: Topology graph information via subgraph and singleton counts at varying threshold ($x$-axis), from 6 to 20 language groups ($y$-axis), with a total of 95 languages. \textbf{Takeaway}: Cross-lingual transferability among similar languages impact memorization.
   }
    \label{fig:intra_cross_topo_results}
\end{figure}
Leveraging the constructed language graph, we measure the topology-based correlation for both intra-topo and cross-topo at various $\theta$. As shown in Fig.~\ref{fig:intra_cross_topo_results}, among cross-topo languages, EM and RM become largely uncorrelated with token counts, spanning from $-0.2$ to $0.05$ with the growing number of language groups. 
While PM has a stronger negative correlation, the correlation becomes generally weaker as more cross-topo language groups are created, from $-0.6$ to $-0.4$. 
This highlights that, \textit{across distinctive language groups, the correlation between memorization and data volume becomes weaker}.
In contrast, consistent with previous findings (cf. Section~\ref{sec:result_llm_mem}), intra-topo $\rho_G$ values grow increasing negative (down to $-0.6$) across memorization metrics, as more similar languages are grouped together (as $\theta$ becomes higher), indicating \textit{an inverse relationship between training data and memorization within similar languages}.

From both cross-topo and intra-topo perspectives, our results show that as MLLMs are trained with richer data from similar languages, memorization decreases —— evidence that 
\textit{cross-lingual transferability} among similar languages plays an essential role in memorization in MLLMs.


\begin{table}[t]
\centering
\small
\begin{tabular}{@{}c|cc|cc@{}}
\toprule
\textbf{Mem. Metric} & \multicolumn{2}{c|}{\textbf{EM}} & \multicolumn{2}{c}{\textbf{PM}} \\
\midrule
\textbf{Model}& $r$ & {$\rho_{G}$}  & $r$ & {$\rho_{G}$}  \\
\midrule
\textsc{mT5-small}  & 0.05 & 0.12 & 0.10 & -0.53 \\
\textsc{mT5-base}   & -0.12 & -0.15 & 0.48 & 0.03 \\
\textsc{mT5-large}  & 0.01 & 0.08 & 0.47 & 0.19 \\
\midrule
\textsc{mGPT-1.3B}  & 0.22 & \textbf{-0.49} & -0.39 & -0.63 \\
\textsc{mGPT-13B}   & 0.18 & -0.13 & -0.39 & \textbf{-0.78} \\
\bottomrule
\end{tabular}
\caption{Correlations between Memorization Rates and Training Data across Models and Scales, with specific $\theta$
for Intra-Topology Correlation. \textbf{Takeaway:} In contrast to $r$, $\rho_G$ presents stronger correlations and more consistent alignment with prior memorization analyses. \textbf{Bold} values indicate the highest-magnitude correlation.
}
\vspace{-5pt}
\label{tab:model-mt5-mgpt}
\end{table}

\begin{table}[t]
\centering
\small
\begin{tabular}{@{}c|cc|cc@{}}
\toprule
\textbf{Mem. Metric} & \multicolumn{2}{c|}{\textbf{RM (BLEU)}} & \multicolumn{2}{c}{\textbf{RM (Rouge-L)}} \\
\midrule
\textbf{Model}& $r$ & {$\rho_{G}$}  & $r$ & {$\rho_{G}$}  \\
\midrule
\textsc{mGPT-1.3B}  & -0.18 & -0.53 & 0.42 & -0.42 \\
\textsc{mGPT-13B}   & -0.21 & -0.31 & 0.4 & -0.04 \\
\bottomrule
\end{tabular}
\caption{Relaxed Memorization Rates across \textsc{mGPT} models, with specific $\theta$ for Intra-Topology Correlation.}
\vspace{-5pt}
\label{tab:mgpt-mem-relaxed}
\end{table}

\subsection{Memorization across Model Architectures \& Scales~\label{results:architecture_scales}}
Since language similarity is model-specific, its scores exhibit different distributions across models. 
We select the specific threshold $\theta$ to better incorporate structural patterns based on language similarity, while confirming that the observed trends hold across a range of thresholds (cf. Appendix~\ref{app:threshold_analysis}).

Table~\ref{tab:model-mt5-mgpt} presents the intra-topology correlations with model-specific thresholds.
For \textsc{mGPT-1.3B} and \textsc{mGPT-13B} - trained on a corpus with a less pronounced long-tailed distribution, $r$ appears positive, seemingly contradictory to previous findings (cf. Sections~\ref{sec:result_llm_mem};~\ref{results:architecture_scales}).
However, leveraging language similarity and filtering out noisy language pairs, $\rho_G$ shows negative correlations, consistent with prior findings.
Notably, with PM, \textsc{mGPT-13B} presents the strongest negative correlation, suggesting that larger models trained on a more balanced corpus reveal the strongest inverse link between memorization and data availability in similar languages. 
In contrary, \textsc{mT5}'s EM results exhibit a different trend compared to \textsc{mGPT} models, which might be attributed to its encoder-decoder architecture. 
As RM is not applicable to \textsc{mt5}-based models (cf. Section~\ref{encoder-decoder-setup}), we show the relaxed memorization metrics for \textsc{mGPT} models in Table~\ref{tab:mgpt-mem-relaxed}. 
We observe a consistent trend aligns with our earlier findings: memorization is negatively correlated with training data quantity among similar languages.




\textbf{In summary}, our analysis and results support the claim that memorization in MLLMs is not shaped solely by training data volume - as commonly observed in computer vision task - but also by intricacies among languages. 
Specifically, when language similarity is incorporated via a topology-based metric, we show that languages with fewer training tokens tend to exhibit higher memorization — a pattern that only becomes evident when language relations are explicitly modeled.

\subsection{Effect of Prompt Length \& Model Scale on Memorization}

\begin{table}[h]
\centering
\footnotesize
\resizebox{\linewidth}{!}{
\begin{tabular}{@{}c|c|cccc@{}}
\toprule
\textbf{Model} & \textbf{Prompt. Len.} & \textbf{EM (\%)} & \textbf{PM} & \textbf{RM (B)} & \textbf{RM (R)} \\
\midrule
\multicolumn{6}{l}{\textit{GPT2 Decoder-only: \textsc{mGPT-101}}} \\
\midrule
\multirow{3}{*}{\textsc{mGPT-101}} & 50  & 0.22 & -44.4 & 3.2 & 9.8 \\
                          & 100  & 0.42 & -41.9 & 3.6 & 10.1 \\
                          & 150 & $\mathbf{0.56}$ & $\mathbf{-40.9}$ & $\mathbf{3.9}$ & $\mathbf{10.1}$ \\
\midrule
\multicolumn{6}{l}{\textit{GPT3 Decoder-only: \textsc{mGPT}-1.3B / 13B}} \\
\midrule
\multirow{3}{*}{\textsc{mGPT-1.3B}} & 50  & 0.31 & -33.7 & $\mathbf{\underline{4.1}}$ & $\mathbf{\underline{5.7}}$  \\
                           & 100  & 0.29 & -32.0 & 3.7 & 5.7  \\
                           & 150 & $\mathbf{0.32}$ &  $\mathbf{-31.1}$ &  3.5 &  4.8  \\
\midrule
\multirow{3}{*}{\textsc{mGPT-13B}}  & 50  & 1.01 & -32.2 & 7.1 & 7.6  \\
                           & 100  & 1.38 & -30.2 & 8.1 & 8.2  \\
                           & 150 & $\mathbf{1.56}$ & $\mathbf{-29.5}$ & $\mathbf{8.6}$ & $\mathbf{8.4}$  \\
\midrule                           
\multicolumn{6}{l}{\textit{Encoder-Decoder: \textsc{mT5} family}} \\
\midrule
\multirow{3}{*}{\textsc{mT5-Small}} & 50  & 0.02 & -66.1  & -- & -- \\
                   & 100  & 0.15 & $\mathbf{\underline{-56.9}}$  & -- & -- \\
                   & 150 & $\mathbf{0.25}$ & -61.3  & -- & -- \\
\midrule
\multirow{3}{*}{\textsc{mT5-Base}} & 50  & 0.07 & -45.7 & -- & --  \\
                   & 100& 0.50 & -35.0 & -- & --  \\
                   & 150 & $\mathbf{0.90}$ & $\mathbf{-31.4}$ & -- & -- \\
\midrule
\multirow{3}{*}{\textsc{mT5-Large}} & 50  & 0.02 & -78.4 & -- & -- \\
                   & 100  & 0.23 & -52.6  & -- & --\\
                   & 150 & $\mathbf{0.49}$ & $\mathbf{-39.0}$ & -- & -- \\

\bottomrule
\end{tabular}}
\caption{
Memorization Rates across various prompt lengths $(35, 85, 135)$, model architectures and scales. The predicted token length is fixed at 15. 
The highest memorization rates for each model are \textbf{bold}.
\textbf{Takeaway}: Overall, the memorization rates increase with the increasing prompt lengths, with a few \underline{exceptions}.
}

\label{tab:all-prompt-scale}

\end{table}

To investigate the effects of experimental setup on memorization, we measure memorization across models of different architectures and scales at varying prompt-length $(35, 85, 135)$, with the fixed output token length of 15. 
The prompt-length refers to prefix-length in the context of decoder-only models.
As shown in Table~\ref{tab:all-prompt-scale},
across all model types, we observe a consistent trend: \textit{longer prompts lead to higher memorization}. 
This pattern holds across the memorization metrics, with a few exceptions, as underlined,
and aligns with prior findings on memorization in monolingual LMs, indicating that longer contexts offer more cues for memorization \cite{carlini2021extracting,carlini2022quantifying}.

In \textsc{GPT-3}–based decoder-only models, we also observe a clear \textbf{scaling effect}: \textit{larger models exhibit stronger memorization}, particularly in exact memorization. For example, EM increases from $0.32\%$ in \textsc{mGPT-1.3B} to $1.56\%$ in \textsc{mGPT-13B} with the prefix of length 135. 
Results in other metrics (e.g., PM, RM) follow this trend with few exceptions. 
In comparison, the encoder-decoder models tells a different story. While memorization generally increases with growing scale (e.g., \textsc{mT5-small} to \textsc{mT5-base}), the largest model (\textsc{mT5-large}) exhibits lower memorization when compared to \textsc{mT5-base}. 

In addition, we observe that \textsc{mT5-large} — without downstream finetuning — produces more broken completions for masked tokens. We hypothesize that this instability may lead to reduced memorization rates in \textsc{mT5-large}, especially in a masked language modeling context. We provide a random example of such unstable generation in Appendix~\ref{app:unstable-generation}.

\subsection{Language-Level Memorization across Prompt Lengths \& Model Scales}
We analyze how language-level memorization varies across different prompt lengths and model scales by computing Pearson correlations of per-language memorization rates under each condition. Across all models, language-level memorization distributions at different prompt lengths remain strongly correlated. For decoder-only models, Pearson correlations consistently exceed $0.9$ in all memorization metrics, while for the 
\textsc{mT5} models, they are generally above $0.8$, with the lowest still above $0.66$. These results indicate that languages with high memorization tend to remain highly memorized regardless of prompt length.  See Table~\ref{tab:empm_prompt_corr_short} and~\ref{tab:rm_prompt_corr_short} for detailed results.

A similar trend holds across model scales. Across all metrics and model scales, the Pearson correlation is consistently shows a strong positive correlation, with the lowest value being $0.71$. These results suggest that memorization tendencies are stable, intrinsic language-level characteristics that generalize across both prompt length and model scale. We observe a ``the poorer get poorer'' phenomenon, where languages with high memorization consistently remain high across settings. See Table~\ref{tab:scale_length_corr_logprob} for full results.

\section{Conclusion and Future Work}
We present the first large-scale study of memorization in MLLMs, grounding observed memorization patterns through language similarity and revealing \textit{cross-linguality} as a key factor shaping memorization in MLLMs.
To this end, we define memorization metrics tailored to language models and 
propose a graph-based correlation measure that incorporates language similarity, uncovering patterns that linear metrics fail to capture. Notably, the tendency for languages with fewer training tokens to exhibit higher memorization, a trend that only becomes apparent when language relationships are explicitly modeled. 
We experiment on a range of language models, across architectures, scales and 95 languages, showing consistent memorization trends. Our findings urge a paradigm shift toward language-aware memorization audits in MLLMs, particularly for under-resourced languages vulnerable to cross-lingual leakage.
We encourage further work at the intersection of multilingualism and memorization to develop effective strategies to mitigate memorization in MLLMs.

\newpage

\section*{Limitations}
Our proposed memorization metric relies on a manually selected similarity threshold to construct the graph, making it sensitive to this parameter and limiting its applicability to languages with low similarity to others, which often become isolated nodes and reduce interpretability. A more robust approach could involve adaptive threshold optimization or the development of threshold-free methods that fully leverage language similarity without requiring manual intervention. 
While our work provides the first large-scale analysis of memorization in MLLMs, we primarily examine models in their pre-trained state and do not explore how fine-tuning or instruction tuning may alter memorization behavior, particularly in task-specific or alignment-sensitive contexts. 
Nonetheless, we believe our study offers a principled and extensible foundation for understanding memorization through the lens of language similarity in multilingual models.

\section*{Ethics Statement}
We comply with the ACL Ethics Policy. This work aims to improve understanding of memorization risks in multilingual language models, with the broader goal of enabling safer and more privacy-preserving NLP systems. All experiments are conducted on publicly available pre-trained models and benchmark datasets. We do not train on, extract from, or attempt to infer sensitive personal information from proprietary or private data.

\section*{Acknowledgements}
XL, YC and JB are funded by the Carlsberg Foundation, under the Semper Ardens: Accelerate programme (project nr. CF21-0454). XL is additionally supported by the EU ChipsJU and the Innovation Fund Denmark through the project CLEVER (no. 101097560). We further acknowledge the support of the AAU AI Cloud and express our gratitude to DeiC for providing computing resources on the LUMI cluster (project nr. DeiC-AAU-N5-2024085-H2-2024-28).
Finally, we thank the Aalborg University AI:X initiative for enabling this work via the AI:SECURITY lab.

\bibliography{custom}

@article{kim2023propile,
  title={Propile: Probing privacy leakage in large language models},
  author={Kim, Siwon and Yun, Sangdoo and Lee, Hwaran and Gubri, Martin and Yoon, Sungroh and Oh, Seong Joon},
  journal={Advances in Neural Information Processing Systems},
  volume={36},
  pages={20750--20762},
  year={2023}
}

@inproceedings{chang2021privacy,
  title={On the privacy risks of algorithmic fairness},
  author={Chang, Hongyan and Shokri, Reza},
  booktitle=EuroSP,
  pages={292--303},
  year={2021},
  organization={IEEE}
}

@inproceedings{carlini2021extracting,
  title={Extracting training data from large language models},
  author={Carlini, Nicholas and Tramer, Florian and Wallace, Eric and Jagielski, Matthew and Herbert-Voss, Ariel and Lee, Katherine and Roberts, Adam and Brown, Tom and Song, Dawn and Erlingsson, Ulfar and others},
  booktitle={30th USENIX security symposium (USENIX Security 21)},
  pages={2633--2650},
  year={2021}
}

@inproceedings{carlini2022quantifying,
  title={Quantifying memorization across neural language models},
  author={Carlini, Nicholas and Ippolito, Daphne and Jagielski, Matthew and Lee, Katherine and Tramer, Florian and Zhang, Chiyuan},
  booktitle={The Eleventh International Conference on Learning Representations},
  year={2022}
}

@article{feldman2020neural,
  title={What neural networks memorize and why: Discovering the long tail via influence estimation},
  author={Feldman, Vitaly and Zhang, Chiyuan},
  journal=NIPS,
  volume={33},
  pages={2881--2891},
  year={2020}
}

@article{garg2023memorization,
  title={Memorization through the lens of curvature of loss function around samples},
  author={Garg, Isha and Ravikumar, Deepak and Roy, Kaushik},
  journal={arXiv preprint arXiv:2307.05831},
  year={2023}
}

@inproceedings{carlini2019secret,
  title={The secret sharer: Evaluating and testing unintended memorization in neural networks},
  author={Carlini, Nicholas and Liu, Chang and Erlingsson, {\'U}lfar and Kos, Jernej and Song, Dawn},
  booktitle={28th USENIX security symposium (USENIX security 19)},
  pages={267--284},
  year={2019}
}

@article{jiang2020exploring,
  title={Exploring the memorization-generalization continuum in deep learning},
  author={Jiang, Ziheng and Zhang, Chiyuan and Talwar, Kunal and Mozer, Michael C},
  journal={arXiv preprint arXiv:2002.03206},
  year={2020}
}

@article{nasr2023scalable,
  title={Scalable extraction of training data from (production) language models},
  author={Nasr, Milad and Carlini, Nicholas and Hayase, Jonathan and Jagielski, Matthew and Cooper, A Feder and Ippolito, Daphne and Choquette-Choo, Christopher A and Wallace, Eric and Tram{\`e}r, Florian and Lee, Katherine},
  journal={arXiv preprint arXiv:2311.17035},
  year={2023}
}

@article{lee2021deduplicating,
  title={Deduplicating training data makes language models better},
  author={Lee, Katherine and Ippolito, Daphne and Nystrom, Andrew and Zhang, Chiyuan and Eck, Douglas and Callison-Burch, Chris and Carlini, Nicholas},
  journal={arXiv preprint arXiv:2107.06499},
  year={2021}
}

@article{xue2020mt5,
  title={mT5: A massively multilingual pre-trained text-to-text transformer},
  author={Xue, Linting and Constant, Noah and Roberts, Adam and Kale, Mihir and Al-Rfou, Rami and Siddhant, Aditya and Barua, Aditya and Raffel, Colin},
  journal={arXiv preprint arXiv:2010.11934},
  year={2020}
}

@article{raffel2020exploring,
  title={Exploring the limits of transfer learning with a unified text-to-text transformer},
  author={Raffel, Colin and Shazeer, Noam and Roberts, Adam and Lee, Katherine and Narang, Sharan and Matena, Michael and Zhou, Yanqi and Li, Wei and Liu, Peter J},
  journal={Journal of machine learning research},
  volume={21},
  number={140},
  pages={1--67},
  year={2020}
}

@misc{cld3,
  title        = {Compact Language Detector v3 (CLD3)},
  howpublished = {\url{https://github.com/google/cld3}},
  note         = {Accessed: 2025-04-30}
}

@inproceedings{littell2017uriel,
  title={Uriel and lang2vec: Representing languages as typological, geographical, and phylogenetic vectors},
  author={Littell, Patrick and Mortensen, David R and Lin, Ke and Kairis, Katherine and Turner, Carlisle and Levin, Lori},
  booktitle={Proceedings of the 15th Conference of the European Chapter of the Association for Computational Linguistics: Volume 2, Short Papers},
  volume={2},
  pages={8--14},
  year={2017}
}

@article{li2025trustworthy,
  title={Trustworthy Machine Learning via Memorization and the Granular Long-Tail: A Survey on Interactions, Tradeoffs, and Beyond},
  author={Li, Qiongxiu and Luo, Xiaoyu and Chen, Yiyi and Bjerva, Johannes},
  journal={arXiv preprint arXiv:2503.07501},
  year={2025}
}

@inproceedings{conneau-etal-2020-unsupervised,
    title = "Unsupervised Cross-lingual Representation Learning at Scale",
    author = "Conneau, Alexis  and
      Khandelwal, Kartikay  and
      Goyal, Naman  and
      Chaudhary, Vishrav  and
      Wenzek, Guillaume  and
      Guzm{\'a}n, Francisco  and
      Grave, Edouard  and
      Ott, Myle  and
      Zettlemoyer, Luke  and
      Stoyanov, Veselin",
    editor = "Jurafsky, Dan  and
      Chai, Joyce  and
      Schluter, Natalie  and
      Tetreault, Joel",
    booktitle = "Proceedings of the 58th Annual Meeting of the Association for Computational Linguistics",
    year = "2020",
    pages = "8440--8451"
}

@article{choi2021analyzingzeroshotcrosslingualtransfer,
  title={Analyzing Zero-shot Cross-lingual Transfer in Supervised NLP Tasks},
  author={Choi, Hyunjin and Kim, Judong and Joe, Seongho and Min, Seungjai and Gwon, Youngjune},
  journal={arXiv preprint arXiv:2101.10649},
  year={2021}
}

@article{pikuliak2021cross,
  title={Cross-lingual learning for text processing: A survey},
  author={Pikuliak, Mat{\'u}{\v{s}} and {\v{S}}imko, Mari{\'a}n and Bielikov{\'a}, M{\'a}ria},
  journal={Expert Systems with Applications},
  volume={165},
  pages={113765},
  year={2021},
  publisher={Elsevier}
}

@article{li2024privacy,
  title={On the privacy effect of data enhancement via the lens of memorization},
  author={Li, Xiao and Li, Qiongxiu and Hu, Zhanhao and Hu, Xiaolin},
  journal=TIFS,
  year={2024},
}

@inproceedings{zhou2004regularization,
  title={A regularization framework for learning from graph data},
  author={Zhou, Dengyong and Sch{\"o}lkopf, Bernhard},
  booktitle={ICML 2004 workshop on statistical relational learning and its connections to other fields (SRL 2004)},
  pages={132--137},
  year={2004}
}

@article{xie2024discovering,
  title={Discovering low-rank subspaces for language-agnostic multilingual representations},
  author={Xie, Zhihui and Zhao, Handong and Yu, Tong and Li, Shuai},
  journal={arXiv preprint arXiv:2401.05792},
  year={2024}
}

@article{gao2023not,
  title={Not all samples are born equal: Towards effective clean-label backdoor attacks},
  author={Gao, Yinghua and Li, Yiming and Zhu, Linghui and Wu, Dongxian and Jiang, Yong and Xia, Shu-Tao},
  journal={Pattern Recognition},
  volume={139},
  pages={109512},
  year={2023},
  publisher={Elsevier}
}

@inproceedings{tramer2022truth,
  title={Truth serum: Poisoning machine learning models to reveal their secrets},
  author={Tram{\`e}r, Florian and Shokri, Reza and San Joaquin, Ayrton and Le, Hoang and Jagielski, Matthew and Hong, Sanghyun and Carlini, Nicholas},
  booktitle=CCS,
  pages={2779--2792},
  year={2022}
}

@inproceedings{chang2023speak,
  title={Speak, Memory: An Archaeology of Books Known to ChatGPT/GPT-4},
  author={Chang, Kent and Cramer, Mackenzie and Soni, Sandeep and Bamman, David},
  booktitle={Proceedings of the 2023 Conference on Empirical Methods in Natural Language Processing},
  pages={7312--7327},
  year={2023}
}

@inproceedings{staabbeyond,
  title={Beyond Memorization: Violating Privacy via Inference with Large Language Models},
  author={Staab, Robin and Vero, Mark and Balunovic, Mislav and Vechev, Martin},
  booktitle={The Twelfth International Conference on Learning Representations}
}

@inproceedings{papineni2002bleu,
  title={Bleu: a method for automatic evaluation of machine translation},
  author={Papineni, Kishore and Roukos, Salim and Ward, Todd and Zhu, Wei-Jing},
  booktitle={Proceedings of the 40th annual meeting of the Association for Computational Linguistics},
  pages={311--318},
  year={2002}
}

@inproceedings{Lauscher2020FromZT,
  title={From Zero to Hero: On the Limitations of Zero-Shot Language Transfer with Multilingual Transformers},
  author={Anne Lauscher and Vinit Ravishankar and Ivan Vulic and Goran Glavas},
  booktitle={Conference on Empirical Methods in Natural Language Processing},
  year={2020},
  url={https://api.semanticscholar.org/CorpusID:226262344}
}

@article{mikolov2013efficient,
  title={Efficient estimation of word representations in vector space},
  author={Mikolov, Tomas and Chen, Kai and Corrado, Greg and Dean, Jeffrey},
  journal={arXiv preprint arXiv:1301.3781},
  year={2013}
}

@article{ammar2016many,
  title={Many languages, one parser},
  author={Ammar, Waleed and Mulcaire, George and Ballesteros, Miguel and Dyer, Chris and Smith, Noah A},
  journal={Transactions of the Association for Computational Linguistics},
  volume={4},
  pages={431--444},
  year={2016},
  publisher={MIT Press One Rogers Street, Cambridge, MA 02142-1209, USA journals-info~…}
}

@inproceedings{vulic-etal-2019-really,
    title = "Do We Really Need Fully Unsupervised Cross-Lingual Embeddings?",
    author = "Vuli{\'c}, Ivan  and
      Glava{\v{s}}, Goran  and
      Reichart, Roi  and
      Korhonen, Anna",
    editor = "Inui, Kentaro  and
      Jiang, Jing  and
      Ng, Vincent  and
      Wan, Xiaojun",
    booktitle = "Proceedings of the 2019 Conference on Empirical Methods in Natural Language Processing and the 9th International Joint Conference on Natural Language Processing (EMNLP-IJCNLP)",
    month = nov,
    year = "2019",
    address = "Hong Kong, China",
    publisher = "Association for Computational Linguistics",
    url = "https://aclanthology.org/D19-1449/",
    doi = "10.18653/v1/D19-1449",
    pages = "4407--4418"
}

@article{conneau2019unsupervised,
  title={Unsupervised cross-lingual representation learning at scale},
  author={Conneau, Alexis and Khandelwal, Kartikay and Goyal, Naman and Chaudhary, Vishrav and Wenzek, Guillaume and Guzm{\'a}n, Francisco and Grave, Edouard and Ott, Myle and Zettlemoyer, Luke and Stoyanov, Veselin},
  journal={arXiv preprint arXiv:1911.02116},
  year={2019}
}

@inproceedings{devlin-etal-2019-bert,
    title = "{BERT}: Pre-training of Deep Bidirectional Transformers for Language Understanding",
    author = "Devlin, Jacob  and
      Chang, Ming-Wei  and
      Lee, Kenton  and
      Toutanova, Kristina",
    editor = "Burstein, Jill  and
      Doran, Christy  and
      Solorio, Thamar",
    booktitle = "Proceedings of the 2019 Conference of the North {A}merican Chapter of the Association for Computational Linguistics: Human Language Technologies, Volume 1 (Long and Short Papers)",
    month = jun,
    year = "2019",
    address = "Minneapolis, Minnesota",
    publisher = "Association for Computational Linguistics",
    url = "https://aclanthology.org/N19-1423/",
    doi = "10.18653/v1/N19-1423",
    pages = "4171--4186"
}

@article{choenni2023languages,
  title={How do languages influence each other? studying cross-lingual data sharing during LM fine-tuning},
  author={Choenni, Rochelle and Garrette, Dan and Shutova, Ekaterina},
  journal={arXiv preprint arXiv:2305.13286},
  year={2023}
}

@inproceedings{Toossi2024ARS,
  title={A Reproducibility Study on Quantifying Language Similarity: The Impact of Missing Values in the URIEL Knowledge Base},
  author={Hasti Toossi and Guo Qing Huai and Jinyu Liu and Eric Khiu and A. Seza Doğru{\"o}z and En-Shiun Annie Lee},
  booktitle={North American Chapter of the Association for Computational Linguistics},
  year={2024},
  url={https://api.semanticscholar.org/CorpusID:269921184}
}

@inproceedings{khan-etal-2025-uriel,
    title = "{URIEL}+: Enhancing Linguistic Inclusion and Usability in a Typological and Multilingual Knowledge Base",
    author = {Khan, Aditya  and
      Shipton, Mason  and
      Anugraha, David  and
      Duan, Kaiyao  and
      Hoang, Phuong H.  and
      Khiu, Eric  and
      Do{\u{g}}ru{\"o}z, A. Seza  and
      Lee, En-Shiun Annie},
    editor = "Rambow, Owen  and
      Wanner, Leo  and
      Apidianaki, Marianna  and
      Al-Khalifa, Hend  and
      Eugenio, Barbara Di  and
      Schockaert, Steven",
    booktitle = "Proceedings of the 31st International Conference on Computational Linguistics",
    month = jan,
    year = "2025",
    address = "Abu Dhabi, UAE",
    publisher = "Association for Computational Linguistics",
    url = "https://aclanthology.org/2025.coling-main.463/",
    pages = "6937--6952"
}

@article{blaschke2025analyzing,
  title={Analyzing the Effect of Linguistic Similarity on Cross-Lingual Transfer: Tasks and Experimental Setups Matter},
  author={Blaschke, Verena and Fedzechkina, Masha and ter Hoeve, Maartje},
  journal={arXiv preprint arXiv:2501.14491},
  year={2025}
}

@inproceedings{de-vries-etal-2022-make,
    title = "Make the Best of Cross-lingual Transfer: Evidence from {POS} Tagging with over 100 Languages",
    author = "de Vries, Wietse  and
      Wieling, Martijn  and
      Nissim, Malvina",
    editor = "Muresan, Smaranda  and
      Nakov, Preslav  and
      Villavicencio, Aline",
    booktitle = "Proceedings of the 60th Annual Meeting of the Association for Computational Linguistics (Volume 1: Long Papers)",
    month = may,
    year = "2022",
    address = "Dublin, Ireland",
    publisher = "Association for Computational Linguistics",
    url = "https://aclanthology.org/2022.acl-long.529/",
    doi = "10.18653/v1/2022.acl-long.529",
    pages = "7676--7685"
}

@inproceedings{adelani-etal-2022-masakhaner,
    title = "{M}asakha{NER} 2.0: {A}frica-centric Transfer Learning for Named Entity Recognition",
    author = "Adelani, David Ifeoluwa  and
      Neubig, Graham  and
      Ruder, Sebastian  and
      Rijhwani, Shruti  and
      Beukman, Michael  and
      Palen-Michel, Chester  and
      Lignos, Constantine  and
      Alabi, Jesujoba O.  and
      Muhammad, Shamsuddeen H.  and
      Nabende, Peter  and
      Dione, Cheikh M. Bamba  and
      Bukula, Andiswa  and
      Mabuya, Rooweither  and
      Dossou, Bonaventure F. P.  and
      Sibanda, Blessing  and
      Buzaaba, Happy  and
      Mukiibi, Jonathan  and
      Kalipe, Godson  and
      Mbaye, Derguene  and
      Taylor, Amelia  and
      Kabore, Fatoumata  and
      Emezue, Chris Chinenye  and
      Aremu, Anuoluwapo  and
      Ogayo, Perez  and
      Gitau, Catherine  and
      Munkoh-Buabeng, Edwin  and
      Memdjokam Koagne, Victoire  and
      Tapo, Allahsera Auguste  and
      Macucwa, Tebogo  and
      Marivate, Vukosi  and
      Mboning, Elvis  and
      Gwadabe, Tajuddeen  and
      Adewumi, Tosin  and
      Ahia, Orevaoghene  and
      Nakatumba-Nabende, Joyce  and
      Mokono, Neo L.  and
      Ezeani, Ignatius  and
      Chukwuneke, Chiamaka  and
      Adeyemi, Mofetoluwa  and
      Hacheme, Gilles Q.  and
      Abdulmumin, Idris  and
      Ogundepo, Odunayo  and
      Yousuf, Oreen  and
      Moteu Ngoli, Tatiana  and
      Klakow, Dietrich",
    editor = "Goldberg, Yoav  and
      Kozareva, Zornitsa  and
      Zhang, Yue",
    booktitle = "Proceedings of the 2022 Conference on Empirical Methods in Natural Language Processing",
    month = dec,
    year = "2022",
    address = "Abu Dhabi, United Arab Emirates",
    publisher = "Association for Computational Linguistics",
    url = "https://aclanthology.org/2022.emnlp-main.298/",
    doi = "10.18653/v1/2022.emnlp-main.298",
    pages = "4488--4508"
}

@article{philippy2023towards,
  title={Towards a common understanding of contributing factors for cross-lingual transfer in multilingual language models: A review},
  author={Philippy, Fred and Guo, Siwen and Haddadan, Shohreh},
  journal={arXiv preprint arXiv:2305.16768},
  year={2023}
}

@inproceedings{chen-etal-2025-large,
    title = "Large Language Models are Easily Confused: A Quantitative Metric, Security Implications and Typological Analysis",
    author = "Chen, Yiyi  and
      Li, Qiongxiu  and
      Biswas, Russa  and
      Bjerva, Johannes",
    editor = "Chiruzzo, Luis  and
      Ritter, Alan  and
      Wang, Lu",
    booktitle = "Findings of the Association for Computational Linguistics: NAACL 2025",
    month = apr,
    year = "2025",
    address = "Albuquerque, New Mexico",
    publisher = "Association for Computational Linguistics",
    url = "https://aclanthology.org/2025.findings-naacl.210/",
    pages = "3810--3827",
    ISBN = "979-8-89176-195-7"
}

@inproceedings{libovicky-etal-2020-language,
    title = "On the Language Neutrality of Pre-trained Multilingual Representations",
    author = "Libovick{\'y}, Jind{\v{r}}ich  and
      Rosa, Rudolf  and
      Fraser, Alexander",
    editor = "Cohn, Trevor  and
      He, Yulan  and
      Liu, Yang",
    booktitle = "Findings of the Association for Computational Linguistics: EMNLP 2020",
    month = nov,
    year = "2020",
    address = "Online",
    publisher = "Association for Computational Linguistics",
    url = "https://aclanthology.org/2020.findings-emnlp.150/",
    doi = "10.18653/v1/2020.findings-emnlp.150",
    pages = "1663--1674"
}

@inproceedings{pires-etal-2019-multilingual,
    title = "How Multilingual is Multilingual {BERT}?",
    author = "Pires, Telmo  and
      Schlinger, Eva  and
      Garrette, Dan",
    editor = "Korhonen, Anna  and
      Traum, David  and
      M{\`a}rquez, Llu{\'i}s",
    booktitle = "Proceedings of the 57th Annual Meeting of the Association for Computational Linguistics",
    month = jul,
    year = "2019",
    address = "Florence, Italy",
    publisher = "Association for Computational Linguistics",
    url = "https://aclanthology.org/P19-1493/",
    doi = "10.18653/v1/P19-1493",
    pages = "4996--5001"
}

@inproceedings{tiyajamorn-etal-2021-language,
    title = "Language-agnostic Representation from Multilingual Sentence Encoders for Cross-lingual Similarity Estimation",
    author = "Tiyajamorn, Nattapong  and
      Kajiwara, Tomoyuki  and
      Arase, Yuki  and
      Onizuka, Makoto",
    editor = "Moens, Marie-Francine  and
      Huang, Xuanjing  and
      Specia, Lucia  and
      Yih, Scott Wen-tau",
    booktitle = "Proceedings of the 2021 Conference on Empirical Methods in Natural Language Processing",
    month = nov,
    year = "2021",
    address = "Online and Punta Cana, Dominican Republic",
    publisher = "Association for Computational Linguistics",
    url = "https://aclanthology.org/2021.emnlp-main.612/",
    doi = "10.18653/v1/2021.emnlp-main.612",
    pages = "7764--7774"
}

@inproceedings{lin-etal-2024-mplm,
    title = "m{PLM}-Sim: Better Cross-Lingual Similarity and Transfer in Multilingual Pretrained Language Models",
    author = "Lin, Peiqin  and
      Hu, Chengzhi  and
      Zhang, Zheyu  and
      Martins, Andre  and
      Schuetze, Hinrich",
    editor = "Graham, Yvette  and
      Purver, Matthew",
    booktitle = "Findings of the Association for Computational Linguistics: EACL 2024",
    month = mar,
    year = "2024",
    address = "St. Julian{'}s, Malta",
    publisher = "Association for Computational Linguistics",
    url = "https://aclanthology.org/2024.findings-eacl.20/",
    pages = "276--310"
}

@article{choenni2022investigating,
  title={Investigating language relationships in multilingual sentence encoders through the lens of linguistic typology},
  author={Choenni, Rochelle and Shutova, Ekaterina},
  journal={Computational Linguistics},
  volume={48},
  number={3},
  pages={635--672},
  year={2022},
  publisher={MIT Press One Broadway, 12th Floor, Cambridge, Massachusetts 02142, USA~…}
}

@article{nllb-24,
    author="{NLLB Team} and Costa-juss{\`a}, Marta R. and Cross, James and {\c{C}}elebi, Onur and Elbayad, Maha and Heafield, Kenneth and Heffernan, Kevin and Kalbassi, Elahe and Lam, Janice and Licht, Daniel and Maillard, Jean and Sun, Anna and Wang, Skyler and Wenzek, Guillaume and Youngblood, Al and Akula, Bapi and Barrault, Loic and Gonzalez, Gabriel Mejia and Hansanti, Prangthip and Hoffman, John and Jarrett, Semarley and Sadagopan, Kaushik Ram and Rowe, Dirk and Spruit, Shannon and Tran, Chau and Andrews, Pierre and Ayan, Necip Fazil and Bhosale, Shruti and Edunov, Sergey and Fan, Angela and Gao, Cynthia and Goswami, Vedanuj and Guzm{\'a}n, Francisco and Koehn, Philipp and Mourachko, Alexandre and Ropers, Christophe and Saleem, Safiyyah and Schwenk, Holger and Wang, Jeff",
    title="Scaling neural machine translation to 200 languages",
    journal="Nature",
    year="2024",
    volume="630",
    number="8018",
    pages="841--846",
    issn="1476-4687",
    doi="10.1038/s41586-024-07335-x",
    url="https://doi.org/10.1038/s41586-024-07335-x"
}

@inproceedings{cavalin-etal-2024-fixing,
    title = "Fixing Rogue Memorization in Many-to-One Multilingual Translators of Extremely-Low-Resource Languages by Rephrasing Training Samples",
    author = "Cavalin, Paulo  and
      Domingues, Pedro Henrique  and
      Pinhanez, Claudio  and
      Nogima, Julio",
    editor = "Duh, Kevin  and
      Gomez, Helena  and
      Bethard, Steven",
    booktitle = "Proceedings of the 2024 Conference of the North American Chapter of the Association for Computational Linguistics: Human Language Technologies (Volume 1: Long Papers)",
    month = jun,
    year = "2024",
    address = "Mexico City, Mexico",
    publisher = "Association for Computational Linguistics",
    url = "https://aclanthology.org/2024.naacl-long.253/",
    doi = "10.18653/v1/2024.naacl-long.253",
    pages = "4503--4514"
}

@inproceedings{lin2004rouge,
  title={Rouge: A package for automatic evaluation of summaries},
  author={Lin, Chin-Yew},
  booktitle={Text summarization branches out},
  pages={74--81},
  year={2004}
}

@article{wichmann2011phonological,
  title={Phonological diversity, word length, and population sizes across languages: The ASJP evidence},
  author={Wichmann, S{\o}ren and Rama, Taraka and Holman, Eric W},
  year={2011},
  publisher={Walter de Gruyter GmbH \& Co. KG}
}

@inproceedings{chen2024multi,
  title={A Multi-Perspective Analysis of Memorization in Large Language Models},
  author={Chen, Bowen and Han, Namgi and Miyao, Yusuke},
  booktitle={Proceedings of the 2024 Conference on Empirical Methods in Natural Language Processing},
  pages={11190--11209},
  year={2024}
}

@inproceedings{liu2024forgetting,
  title={Forgetting Curve: A Reliable Method for Evaluating Memorization Capability for Long-Context Models},
  author={Liu, Xinyu and Zhao, Runsong and Huang, Pengcheng and Xiao, Chunyang and Li, Bei and Wang, Jingang and Xiao, Tong and Zhu, Jingbo},
  booktitle={Proceedings of the 2024 Conference on Empirical Methods in Natural Language Processing},
  pages={4667--4682},
  year={2024}
}

@article{li2024rome,
  title={Rome: Memorization insights from text, logits and representation},
  author={Li, Bo and Zhao, Qinghua and Wen, Lijie},
  journal={arXiv preprint arXiv:2403.00510},
  year={2024}
}

@article{stoehr2024localizing,
  title={Localizing paragraph memorization in language models},
  author={Stoehr, Niklas and Gordon, Mitchell and Zhang, Chiyuan and Lewis, Owen},
  journal={arXiv preprint arXiv:2403.19851},
  year={2024}
}

@article{haviv2022understanding,
  title={Understanding transformer memorization recall through idioms},
  author={Haviv, Adi and Cohen, Ido and Gidron, Jacob and Schuster, Roei and Goldberg, Yoav and Geva, Mor},
  journal={arXiv preprint arXiv:2210.03588},
  year={2022}
}

@article{zhao2024word,
  title={Word Order and World Knowledge},
  author={Zhao, Qinghua and Ravishankar, Vinit and Garneau, Nicolas and S{\o}gaard, Anders},
  journal={arXiv preprint arXiv:2403.00876},
  year={2024}
}

\appendix
\clearpage

\section{Language-specific Subspaces}
\label{sec:langsubspace}
The algorithm for identifying language-specific subspace is as in Algorithm~\ref{algo:lsar}, refer to \citet{xie2024discovering} for more details.

\begin{algorithm}
    \caption{Language-specific Subspace Identification}\label{algo:lsar}
    \textbf{Input}: Languages' mean Embeddings $\mM$, rank of subspace $r$.
    
    \textbf{Output}: Language-agnostic component $\mu$, language-specific subspace $\mM_s$, coordinates $\Gamma$.
    
    /* 1) Approximate $\mM$ in low rank */

    $\mu' \leftarrow \frac{1}{d}\mM \mathbbm{1}^{\intercal}$

    $\mM'_{s, \_, \Gamma} \leftarrow \text{Top-}r \text{SVD}(\mM - \mu'\mathbbm{1}^{\intercal});$

    $\mM' \leftarrow \mu' \mathbbm{1}^{\intercal} + \mM'_{s} \Gamma'^{\intercal}$;

    /* 2) Force orthogonality  */

    $\mu \leftarrow (1/||\mM'+ \mathbbm{1}||^{2}) \mM'^{+} \mathbbm{1} $

    $\mM_s, \_, \Gamma \leftarrow \text{Top-}r \text{SVD}(\mM'-\mu \mathbbm{1}^{\intercal})$

\end{algorithm}

\section{Appendix}
\label{sec:appendix}

\subsection{Models detail}
\begin{table}[h]
\centering
\resizebox{\linewidth}{!}{
\begin{tabular}{l|c|c|c|l}
\toprule
\textbf{Model} & \textbf{\#Params} & \textbf{\#Langs. (used.)} & \textbf{Architecture} & \textbf{Layers} \\
\midrule
\textsc{mGPT-101}      & 560M   & 101 (95)     & GPT-2 based       & 24 \\
\textsc{mGPT-1.3B}     & 1.3B   & 61 (48)     & GPT-3 based       & 24 \\
\textsc{mGPT-13B}     & 13B    & 61 (48)     & GPT-3 based       & 40 \\
\textsc{mT5-SMALL} & 300M   & 101 (95)    & Encoder-Decoder   & 8 \\
\textsc{mT5-BASE}  & 580M   & 101 (95)    & Encoder-Decoder   & 12 \\
\textsc{mT5-LARGE} & 1.2B   & 101 (95)    & Encoder-Decoder   & 24 \\
\bottomrule
\end{tabular}
}
\caption{MLLMs and their Scale, Datasets, Languages (analyzed), Architectures.}
\label{tab:model_selection}
\end{table}

\subsection{Cross-lingual correlation}
\label{app:threshold_analysis}

\begin{table}[h!]
\centering
\resizebox{\linewidth}{!}{
\begin{tabular}{l|c|c|c|c|c|c|c|c}
\toprule
 & \multicolumn{8}{c}{\textbf{Threshold $\theta$}} \\
\midrule
   \textsc{mGPT-101}    & 0.31 & 0.33 & 0.35 & 0.37 & 0.39 & 0.41 & 0.43 & 0.45 \\
\# \textbf{Subgraph}        & 11    & 14    & 16   & 18   & 19   & 25   & 26   & 35 \\
\# \textbf{Single Point}    & 7     & 9     & 10   & 11   & 11   & 18   & 18   & 24 \\
\midrule
EM Intra        & -0.13 & -0.17 & -0.20 & -0.26 & -0.24 & -0.24 & -0.19 & -0.17 \\
EM Cross        & -0.15 & -0.01 & -0.03 & -0.01 & -0.02 &  0.04 &  0.04 & -0.09 \\
\midrule
PM Intra        & -0.31 & -0.33 & -0.38 & -0.43 & -0.51 & -0.56 & -0.54 & -0.57 \\
PM Cross        & -0.53 & -0.46 & -0.42 & -0.42 & -0.45 & -0.35 & -0.36 & -0.36 \\
\midrule
RM (B) Intra    & -0.27 & -0.32 & -0.35 & -0.39 & -0.38 & -0.36 & -0.33 & -0.31 \\
RM (B) Cross    & -0.07 &  0.09 & -0.00 &  0.02 & -0.03 &  0.04 &  0.04 & -0.12 \\
\midrule
RM (R) Intra    & -0.20 & -0.26 & -0.27 & -0.28 & -0.32 & -0.30 & -0.26 & -0.24 \\
RM (R) Cross    &  0.05 &  0.10 &  0.06 &  0.05 &  0.05 &  0.41 &  0.42 &  0.18 \\
\bottomrule
\end{tabular}
}
\caption{Cross-topo vs. intra-topo correlation at low thresholds for mGPT-101.}
\label{tab:mgpt101}
\end{table}

\begin{table}[h!]
\centering
\resizebox{\linewidth}{!}{
\begin{tabular}{l|c|c|c|c|c|c|c|c}
\toprule
 & \multicolumn{8}{c}{\textbf{Threshold $\theta$}} \\
\midrule
   \textsc{mGPT-1.3B}    & 0.82 & 0.83 & 0.84 & 0.85 & 0.86 & 0.87 & 0.88 & 0.89 \\
\# \textbf{Subgraph}        & 8    & 11   & 12   & 13   & 22   & 28   & 31   & 33 \\
\# \textbf{Single Point}    & 5    & 7    & 8    & 9    & 17   & 23   & 26   & 27 \\
\midrule
EM Intra& -0.04 & -0.09 & -0.12 & -0.11 & -0.16 & -0.26 & -0.49 & -0.43 \\
EM Cross& 0.74  & 0.76  & 0.56  & 0.55  & 0.25  & 0.19  & 0.20  & 0.19 \\
\midrule
PM Intra         & -0.45 & -0.46 & -0.49 & -0.51 & -0.59 & -0.50 & -0.60 & -0.63 \\
PM Cross         & 0.05  & 0.18  & 0.18  & 0.14  & -0.06 & -0.13 & -0.20 & -0.22 \\
\midrule
RM (B) Intra       & -0.20 & -0.22 & -0.27 & -0.29 & -0.35 & -0.40 & -0.53 & -0.50 \\
RM (B) Cross       & 0.34  & 0.43  & 0.40  & 0.34  & 0.04  & -0.06 & -0.09 & -0.11 \\
\midrule
RM (R) Intra       & 0.24  & 0.19  & 0.09  & 0.05  & -0.14 & -0.27 & -0.42 & -0.35 \\
RM (R) Cross       & -0.03 & -0.17 & -0.16 & -0.14 & 0.00  & 0.25  & 0.28  & 0.32 \\
\bottomrule
\end{tabular}
}
\caption{Cross-topo vs. intra-topo correlation at high thresholds for mGPT-1.3B.}
\label{tab:mgpt13b}
\end{table}

\begin{table}[h!]
\centering
\resizebox{\linewidth}{!}{
\begin{tabular}{l|cccccccc}
\toprule
 & \multicolumn{8}{c}{\textbf{Threshold $\theta$}} \\
\midrule
\textsc{mGPT-13B}         & 0.28 & 0.30 & 0.32 & 0.34 & 0.36 & 0.38 & 0.40 & 0.42 \\
\# \textbf{Subgraph}      & 4    & 8    & 10   & 14   & 22   & 26   & 31   & 31 \\
\# \textbf{Single Point}  & 2    & 5    & 7    & 10   & 13   & 17   & 21   & 21 \\
\midrule
EM Intra                  & -0.07 & -0.12 & -0.10 & -0.10 &  0.18 &  0.19 &  0.17 &  0.17 \\
EM Cross                  &  0.15 &  0.29 &  0.23 &  0.11 &  0.37 &  0.31 &  0.30 &  0.30 \\
\midrule
PM Intra                  & -0.35 & -0.42 & -0.46 & -0.55 & -0.57 & -0.65 & -0.78 & -0.78 \\
PM Cross                  & -0.85 &  0.28 &  0.28 &  0.17 & -0.00 & -0.12 & -0.25 & -0.25 \\
\midrule
RM (B) Intra              & -0.21 & -0.27 & -0.27 & -0.31 & -0.18 & -0.21 & -0.33 & -0.33 \\
RM (B) Cross              & -0.96 &  0.26 &  0.27 &  0.14 &  0.19 &  0.08 & -0.07 & -0.07 \\
\midrule
RM (R) Intra              &  0.08 &  0.10 & -0.01 & -0.04 &  0.12 &  0.23 &  0.20 &  0.20 \\
RM (R) Cross              &  0.56 &  0.02 &  0.01 &  0.08 &  0.32 &  0.34 &  0.44 &  0.44 \\
\bottomrule
\end{tabular}
}
\caption{Cross-topology vs.\ intra-topology Pearson correlation at varying thresholds for \textsc{mGPT-13B}.}
\label{tab:topo_corr}
\end{table}

\begin{table}[h!]
\centering
\resizebox{\linewidth}{!}{
\begin{tabular}{l|cccccccc}
\toprule
 & \multicolumn{8}{c}{\textbf{Threshold $\theta$}} \\
\midrule
\textsc{mT5-SMALL}         & 0.54 & 0.56 & 0.58 & 0.60 & 0.62 & 0.64 & 0.66 & 0.68 \\
\# \textbf{Subgraph}       & 30   & 38   & 46   & 52   & 56   & 66   & 72   & 77 \\
\# \textbf{Single Point}   & 16   & 23   & 29   & 36   & 41   & 57   & 61   & 70 \\
\midrule
EM Intra                   & 0.27  & 0.27  & 0.26  & 0.24  & 0.20  & 0.22  & 0.16  & 0.12 \\
EM Cross                   & -0.14 & -0.06 & -0.03 & -0.04 & -0.02 & -0.04 & -0.01 &  0.01 \\
\midrule
PM Intra                   & -0.13 & -0.13 & -0.11 & -0.12 & -0.19 & -0.32 & -0.38 & -0.53 \\
PM Cross                   &  0.33 &  0.15 &  0.17 &  0.10 &  0.11 &  0.16 &  0.14 &  0.12 \\
\bottomrule
\end{tabular}
}
\caption{Cross-topology vs.\ intra-topology Pearson correlation at varying thresholds for \textsc{mT5-Small}.}
\label{tab:topo_corr_mt5small}
\end{table}

\begin{table}[!h]
\centering
\resizebox{\linewidth}{!}{
\begin{tabular}{l|cccccccc}
\toprule
 & \multicolumn{8}{c}{\textbf{Threshold $\theta$}} \\
\midrule
\textsc{mT5-BASE}          & 0.72 & 0.74 & 0.76 & 0.78 & 0.80 & 0.82 & 0.84 & 0.86 \\
\# \textbf{Subgraph}       & 1    & 2    & 7    & 14   & 29   & 48   & 62   & 74 \\
\# \textbf{Single Point}   & 0    & 0    & 6    & 9    & 21   & 39   & 50   & 64 \\
\midrule
EM Intra                   & -0.15 & -0.13 & -0.10 & -0.12 &  0.04 &  0.04 & -0.02 & -0.14 \\
EM Cross                   &  0.00 & -1.00 & -0.24 & -0.27 & -0.07 & -0.14 & -0.20 & -0.16 \\
\midrule
PM Intra                   &  0.20 &  0.13 &  0.07 &  0.07 &  0.15 &  0.22 &  0.13 &  0.03 \\
PM Cross                   &  0.00 & -1.00 & -0.44 & -0.16 &  0.04 &  0.21 &  0.27 &  0.28 \\
\bottomrule
\end{tabular}
}
\caption{Cross-topology vs.\ intra-topology Pearson correlation at varying thresholds for \textsc{mT5-Base}.}
\label{tab:topo_corr_mt5base}
\end{table}

\begin{table}[!h]
\centering
\resizebox{\linewidth}{!}{
\begin{tabular}{l|cccccccc}
\toprule
 & \multicolumn{8}{c}{\textbf{Threshold $\theta$}} \\
\midrule
\textsc{mT5-Large}         & 0.85 & 0.86 & 0.87 & 0.88 & 0.89 & 0.90 & 0.91 & 0.92 \\
\# \textbf{Subgraph}       & 7    & 9    & 21   & 27   & 51   & 65   & 79   & 86 \\
\# \textbf{Single Point}   & 6    & 8    & 18   & 23   & 48   & 60   & 74   & 83 \\
\midrule
EM Intra                   & 0.18 & 0.24 & 0.19 & 0.17 & 0.27 & 0.27 & 0.18 & 0.07 \\
EM Cross                   & -0.31 & -0.22 & -0.11 & -0.14 & 0.08 & -0.06 & -0.00 & 0.02 \\
\midrule
PM Intra                   & 0.30 & 0.27 & 0.21 & 0.19 & 0.20 & 0.23 & 0.12 & -0.13 \\
PM Cross                   & 0.43 & 0.52 & 0.52 & 0.52 & 0.51 & 0.57 & 0.52 & 0.50 \\
\bottomrule
\end{tabular}
}
\caption{Cross-topology vs.\ intra-topology Pearson correlation at varying thresholds for \textsc{mT5-Large}.}
\label{tab:topo_corr_mt5large}
\end{table}

\clearpage

\subsection{Prompt length impact}
\label{sec:cross}
\begin{table}[!h]
\centering
\footnotesize
\resizebox{\linewidth}{!}{
\begin{tabular}{@{}l|cc|cc@{}}
\toprule
\textbf{Model} & \multicolumn{2}{c|}{\textbf{EM}} & \multicolumn{2}{c}{\textbf{PM}} \\
\cmidrule(lr){2-3} \cmidrule(lr){4-5}
& 50 vs. 100 & 100 vs. 150 & 50 vs. 100 & 100 vs. 150 \\
\midrule
\multicolumn{5}{l}{\textit{GPT2 Decoder-only: \textsc{mGPT-101}}} \\
\midrule
\textsc{mGPT-101}  & 0.97 & 0.98 & 0.99 & 0.99 \\
\midrule
\multicolumn{5}{l}{\textit{GPT3 Decoder-only: \textsc{mGPT-1.3B / 13B}}} \\
\midrule
\textsc{mGPT-1.3B} & 0.90 & 0.98 & 0.99 & 0.99 \\
\textsc{mGPT-13B}  & 0.96 & 0.99 & 0.99 & 0.99 \\
\midrule
\multicolumn{5}{l}{\textit{Encoder-Decoder: \textsc{mT5} family}} \\
\midrule
\textsc{mT5-Small} & 0.81 & 0.96 & 0.84 & 0.88 \\
\textsc{mT5-Base}  & 0.86 & 0.97 & 0.99 & 0.98 \\
\textsc{mT5-Large} & 0.66 & 0.94 & 0.94 & 0.97 \\
\bottomrule
\end{tabular}
}
\caption{
Correlation of memorization metrics (exact and family vs probability) between prompt lengths 50 vs 100 and 100 vs 150 across model families. 
EM = Exact Memorization, PM = Probability Memorization.
}
\label{tab:empm_prompt_corr_short}
\end{table}

\begin{table}[!h]
\centering
\footnotesize
\resizebox{\linewidth}{!}{
\begin{tabular}{@{}l|cc|cc@{}}
\toprule
\textbf{Model} & \multicolumn{2}{c|}{\textbf{RM BLEU}} & \multicolumn{2}{c}{\textbf{RM ROUGE-L}} \\
\cmidrule(lr){2-3} \cmidrule(lr){4-5}
& 50 vs. 100 & 100 vs. 150 & 50 vs. 100 & 100 vs. 150 \\
\midrule
\multicolumn{5}{l}{\textit{GPT2 Decoder-only: \textsc{mGPT-101}}} \\
\midrule
\textsc{mGPT-101}  & 0.92 & 0.97 & 0.99 & 0.99 \\
\midrule
\multicolumn{5}{l}{\textit{GPT3 Decoder-only: \textsc{mGPT-1.3B / 13B}}} \\
\midrule
\textsc{mGPT-1.3B} & 0.95 & 0.99 & 0.99 & 0.99 \\
\textsc{mGPT-13B}  & 0.99 & 0.99 & 0.99 & 0.99 \\
\midrule
\end{tabular}
}
\caption{
Correlation of relaxed memorization metrics between different prompt lengths.
}
\label{tab:rm_prompt_corr_short}
\end{table}

\begin{table}[!h]
\centering
\resizebox{\linewidth}{!}{
\begin{tabular}{l|ccc}
\toprule
\textbf{Model Pair} & \textbf{Mem. Metric} & $r$ \\
\midrule
\multirow{2}{*}{\textsc{mT5-Small} vs.\ \textsc{mT5-Base}}        
  & EM & 0.71  \\
  & PM & 0.76  \\
\midrule
\multirow{2}{*}{\textsc{mT5-Base} vs.\ \textsc{mT5-Large}}        
  & EM & 0.81 \\
  & PM & 0.72 \\
\midrule
\multirow{4}{*}{\textsc{mGPT-1.3B} vs.\ \textsc{mGPT-13B}}        
  & EM & 0.92 \\
  & PM & 0.99 \\
  & RM (BLEU) & 0.97 \\
  & RM (ROUGE-L) & 0.99 \\
\bottomrule
\end{tabular}
}
\caption{Pairwise memorization correlation ($r$) between adjacent model scales for exact memorization (EM), probability memorization (PM), and reference match metrics (RM).}
\label{tab:scale_length_corr_logprob}
\end{table}

\subsection{Layer-wise Lang2Vec correlation~\label{app:layer-correlation}}

We include supplementary visualizations showing how various linguistic feature correlations evolve across layers for different multilingual models.

\begin{figure}[!h]
    \centering
    \includegraphics[width=\linewidth]{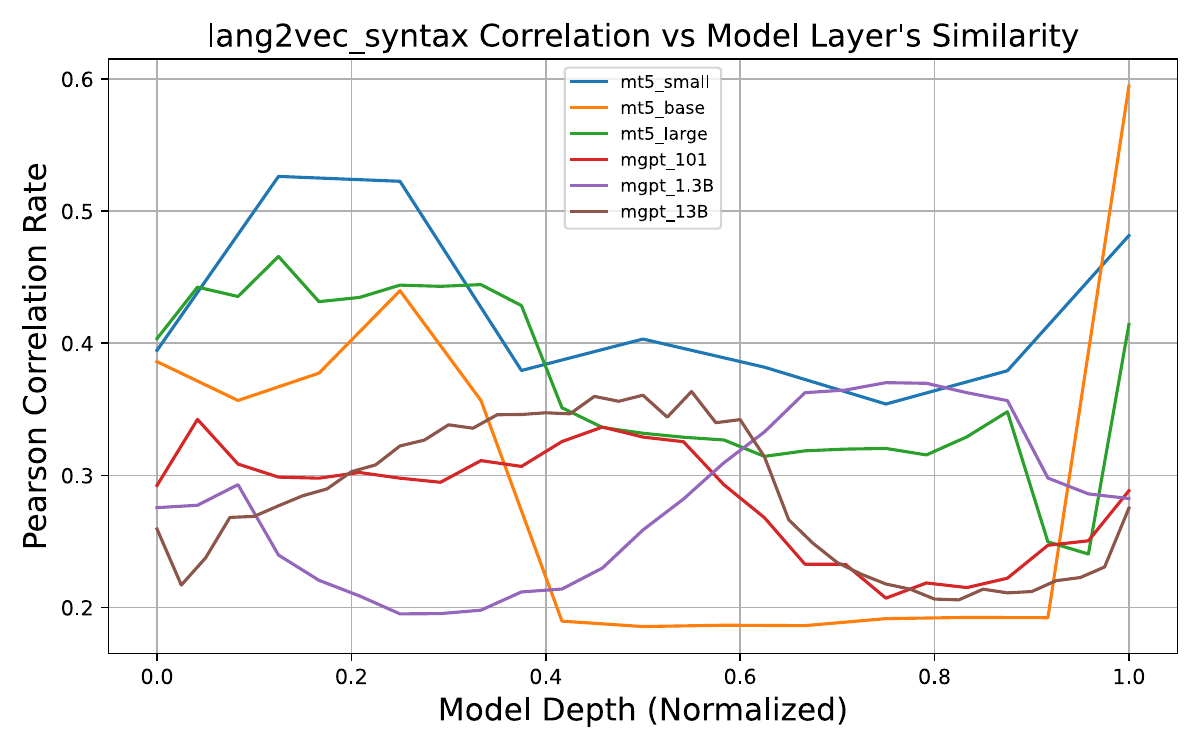}
    \caption{Layer-wise trend for \texttt{Lang2Vec (Syntax)}.}
    \label{fig:lang2vec_syntax}
\end{figure}

\begin{figure}[!h]
    \centering
    \includegraphics[width=\linewidth]{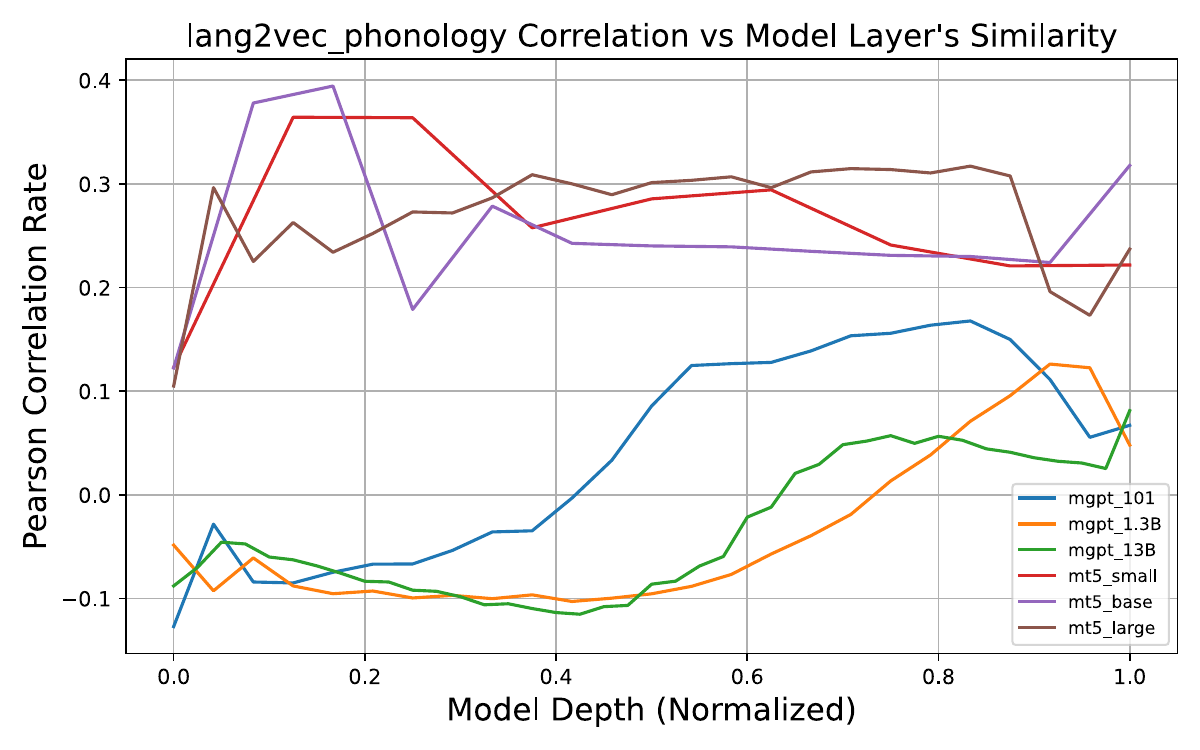}
    \caption{Layer-wise trend for \texttt{Lang2Vec (Phonology)}.}
    \label{fig:lang2vec_phonology}
\end{figure}

\begin{figure}[!h]
    \centering
    \includegraphics[width=\linewidth]{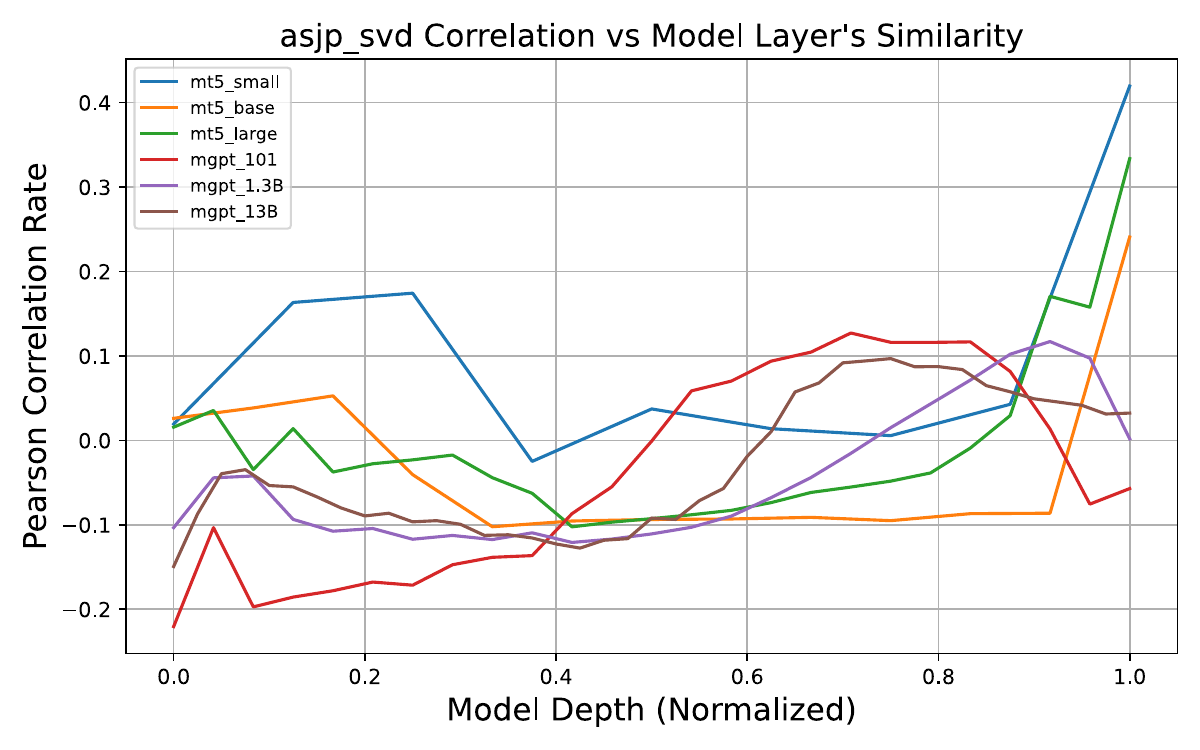}
    \caption{Layer-wise trend for \texttt{ASJP (SVD)}.}
    \label{fig:asjp_svd}
\end{figure}

\begin{figure}[!h]
    \centering
    \includegraphics[width=\linewidth]{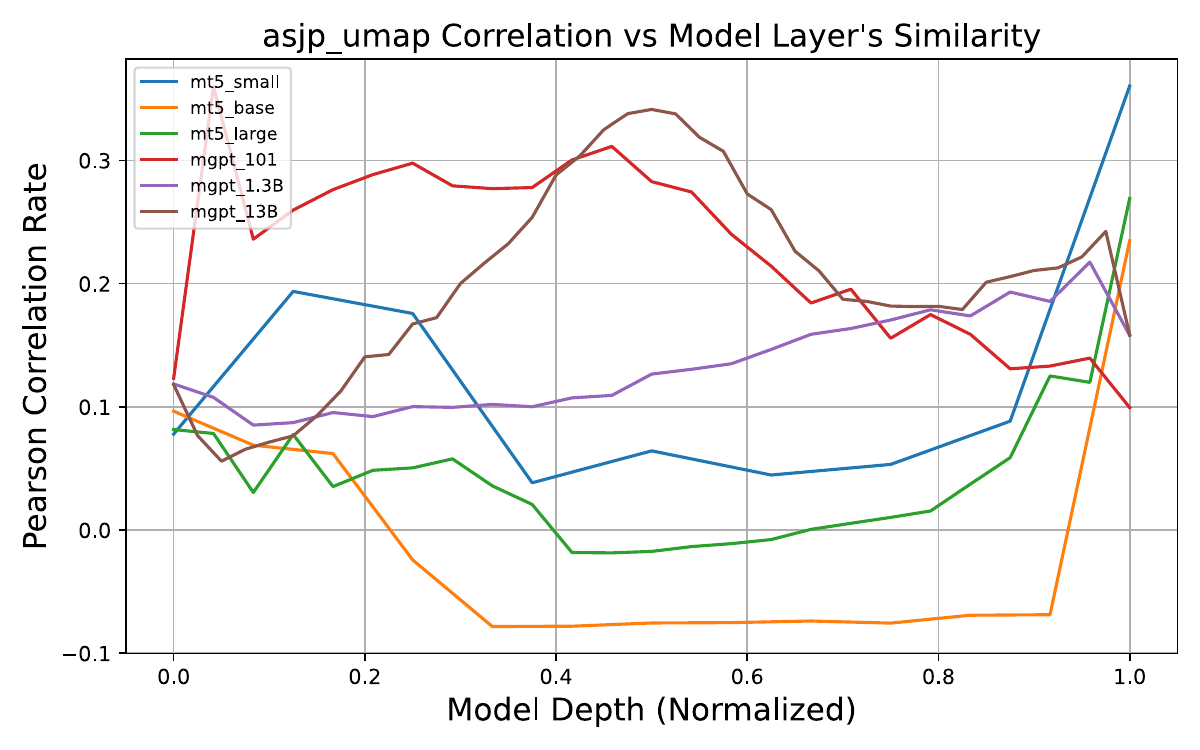}
    \caption{Layer-wise trend for \texttt{ASJP (UMAP)}.}
    \label{fig:asjp_umap}
\end{figure}

\subsection{Examples of Exact Memorization}
\begin{tcolorbox}[
  title=Danish \textsc{mGPT-101},
  colback=gray!5,
  colframe=gray!30,
  fonttitle=\bfseries,
  sharp corners,
  enhanced,
  breakable,
  width=\columnwidth
]
\footnotesize
\textbf{Prompt:}

\ttfamily
Americas Best Value Inn Santa Rosa tilbyder også mange faciliteter der vil berige dit ophold i Santa Rosa (CA). Hotellet tilbyder sine gæster adgang til et stort udvalg af servicetilbud, som trådløst internet i fællesområder, parkering, familieværelse. Hotellets bekvemmeligheder er særligt udvalgt for at sikre den højeste komfort. På nogle af værelserne kan gæsterne finde internetadgang - trådløst,

\vspace{0.5em}
\textbf{Reference:}

ikke-rygerværelser, aircondition, skrivebord

\vspace{0.5em}
\textbf{Prediction:}

ikke-rygerværelser, aircondition, skrivebord
\end{tcolorbox}

\begin{tcolorbox}[
  title=Danish \textsc{mT5-BASE},
  colback=gray!5,
  colframe=gray!30,
  fonttitle=\bfseries,
  sharp corners,
  enhanced,
  breakable,
  width=\columnwidth
]
\footnotesize
\textbf{Prompt:}

\ttfamily
- Se på kort Mere om Pensjonat Mi \textbf{\texttt{<extra\_id\_0>}} Miłosna er indrettet til \textbf{\texttt{<extra\_id\_1>}}- og forretningsrejse \textbf{\texttt{<extra\_id\_2>}}er idéelt i Kwidzyn; én af byens mest populære beliggenheder. Herfra har gæster glæde af nem adgang til alt, hvad denne livlige by kan tilbyde. Med sin praktisk \textbf{\texttt{<extra\_id\_3>}} b \textbf{\texttt{<extra\_id\_4>}} dette hotel nem adgang til byens vigtigste sev

\vspace{0.5em}
\textbf{Reference:}

\textbf{\texttt{<extra\_id\_0>}}łosna Pensjonat \textbf{\texttt{<extra\_id\_1>}} både ferie \textbf{\texttt{<extra\_id\_2>}}nde og ligg \textbf{\texttt{<extra\_id\_3>}}e \textbf{\texttt{<extra\_id\_4>}}eliggenhed tilbyder

\vspace{0.5em}
\textbf{Prediction:}

\textbf{\texttt{<extra\_id\_0>}}łosna Pensjonat \textbf{\texttt{<extra\_id\_1>}} både ferie \textbf{\texttt{<extra\_id\_2>}}nde og ligg \textbf{\texttt{<extra\_id\_3>}}e \textbf{\texttt{<extra\_id\_4>}}eliggenhed tilbyder
\end{tcolorbox}

\begin{tcolorbox}[title=German \textsc{mGPT-101}, colback=gray!5, colframe=gray!30, fonttitle=\bfseries, sharp corners, enhanced, breakable]
\footnotesize
\textbf{Prompt:}

\begin{flushleft}
\ttfamily
Wir denken ebenfalls, dass solcherlei akzeptabel recherchierte Tests, überaus hilfreich sind. Trotzdem wollen wir du jene Gattung von Produktvorstellungen nicht anbieten, weil der Markt außerordentlich schnelllebig und dynamisch ist und zum wiederholten Male neumodische Produktkette dazukommen und die "alten" Produktmodelle uninteressant werden, egal um welches Produkt es geht. Deswegen bieten wir auf unserer Seite ausschließlich eine Darstellung von den jetzigen 5 Produkte an. Somit kann
\end{flushleft}

\vspace{0.5em}
\textbf{Reference:}
\begin{flushleft}
\ttfamily
man sich selbsttätig seine Favoriten intuitiv raussuchen
\end{flushleft}

\vspace{0.5em}
\textbf{Prediction:}
\begin{flushleft}
\ttfamily
man sich selbsttätig seine Favoriten intuitiv raussuchen
\end{flushleft}
\end{tcolorbox}

\begin{tcolorbox}[title=German \textsc{mT5-BASE}, colback=gray!5, colframe=gray!30, fonttitle=\bfseries, sharp corners, enhanced, breakable]
\footnotesize
\textbf{Prompt:}

\begin{flushleft}
\ttfamily
die Versandkosten ungeachtet dessen überaus nie \textbf{\texttt{<extra\_id\_0>}}halten werden oder keineswegs erst anfallen. Zu diesem Zweck gehören die Leistung, die getrennten Einstellungen, die Größe des Körpers und der genaue Einsatzbereich. Das \textbf{\texttt{<extra\_id\_1>}} ein außergewöhnlich breites Angebot von Erzeugnissen fix \textbf{\texttt{<extra\_id\_2>}}roduzenten ak \textbf{\texttt{<extra\_id\_3>}}. Häufig werden lediglich wenige be \textbf{\texttt{<extra\_id\_4>}}t, weil die
\end{flushleft}

\vspace{0.5em}
\textbf{Reference:}
\begin{flushleft}
\ttfamily
\textbf{\texttt{<extra\_id\_0>}}drig ge \textbf{\texttt{<extra\_id\_1>}} Kaufportal offeriert \textbf{\texttt{<extra\_id\_2>}} vom P \textbf{\texttt{<extra\_id\_3>}}kurat wie von Händlern \textbf{\texttt{<extra\_id\_4>}}rücksichtig
\end{flushleft}

\vspace{0.5em}
\textbf{Prediction:}
\begin{flushleft}
\ttfamily
\textbf{\texttt{<extra\_id\_0>}}drig ge \textbf{\texttt{<extra\_id\_1>}} Kaufportal offeriert \textbf{\texttt{<extra\_id\_2>}} vom P \textbf{\texttt{<extra\_id\_3>}}kurat wie von Händlern \textbf{\texttt{<extra\_id\_4>}}rücksichtig
\end{flushleft}
\end{tcolorbox}

\begin{tcolorbox}[title=English \textsc{mGPT-101}, colback=gray!5, colframe=gray!30, fonttitle=\bfseries, sharp corners, enhanced, breakable]
\footnotesize
\textbf{Prompt:}

\begin{flushleft}
\ttfamily
exactly dimension of Modern Ideas Sports Wallpapers Backgrounds Hd On The App Store was 246x246 pixels. You can also look for some pictures that related to Modern Ideas Sports Wallpapers Backgrounds Hd On The App Store by scroll down to collection on below this picture. If you want to find the other picture or article about Sports Wallpapers just push the next button or previous button; or if you are interested in similar pictures of Modern Ideas Sports Wallpapers Backgrounds Hd On
\end{flushleft}

\vspace{0.5em}
\textbf{Reference:}
\begin{flushleft}
\ttfamily
The App Store, you are free to browse through search feature that
\end{flushleft}

\vspace{0.5em}
\textbf{Prediction:}
\begin{flushleft}
\ttfamily
The App Store, you are free to browse through search feature that
\end{flushleft}
\end{tcolorbox}

\begin{tcolorbox}[title=English \textsc{mT5-BASE}, colback=gray!5, colframe=gray!30, fonttitle=\bfseries, sharp corners, enhanced, breakable]
\footnotesize
\textbf{Prompt:}

\begin{flushleft}
\ttfamily
the administration announced a \$6 million investment over two years for provider education and outreach. Expand support \textbf{\texttt{<extra\_id\_0>}} with Alzheimer \textbf{\texttt{<extra\_id\_1>}}their families: \textbf{\texttt{<extra\_id\_2>}} with Alzheimer’s disease and their families and care \textbf{\texttt{<extra\_id\_3>}}requires giving them the tools that they need, helping to plan for future needs, and ensuring that safety and dignity are  \textbf{\texttt{<extra\_id\_4>}}ed,” the report says. The announcement proposes an investment
\end{flushleft}

\vspace{0.5em}
\textbf{Reference:}
\begin{flushleft}
\ttfamily
\textbf{\texttt{<extra\_id\_0>}} for people \textbf{\texttt{<extra\_id\_1>}}’s disease and  \textbf{\texttt{<extra\_id\_2>}} “Supporting people \textbf{\texttt{<extra\_id\_3>}}givers  \textbf{\texttt{<extra\_id\_4>}}maintain
\end{flushleft}

\vspace{0.5em}
\textbf{Prediction:}
\begin{flushleft}
\ttfamily
\textbf{\texttt{<extra\_id\_0>}} for people \textbf{\texttt{<extra\_id\_1>}}’s disease and  \textbf{\texttt{<extra\_id\_2>}} “Supporting people \textbf{\texttt{<extra\_id\_3>}}givers  \textbf{\texttt{<extra\_id\_4>}}maintain
\end{flushleft}
\end{tcolorbox}

\begin{CJK}{UTF8}{gbsn}
\begin{tcolorbox}[title=Chinese \textsc{mGPT-101}, colback=gray!5, colframe=gray!30, fonttitle=\bfseries, sharp corners, enhanced, breakable]
\footnotesize
\textbf{Prompt:}

\begin{flushleft}
\ttfamily
大。 新宝gg游戏平台网页版第88届奥斯卡颁奖礼已经落下帷幕,与其有关的话题还在持续。获奖的近20部影片中有不少改编自小说,单是入围“最佳影片”角逐的9部影片就有5部改编自小说。其中,像《荒野猎人》《房间》等获奖影片的原著小说都出版了中文版。此外,获提名的《火星救援》《卡罗尔》等四部电影的原著小说也有了中文版。看过电影后,不妨去读读这些原著小说。 昨日早上5时
\end{flushleft}

\vspace{0.5em}
\textbf{Reference:}
\begin{flushleft}
\ttfamily
许,在距离爆炸现场南侧不到400米处的天津港进口
\end{flushleft}

\vspace{0.5em}
\textbf{Prediction:}
\begin{flushleft}
\ttfamily
许,在距离爆炸现场南侧不到400米处的天津港进口
\end{flushleft}
\end{tcolorbox}
\end{CJK}

\begin{CJK}{UTF8}{gbsn}
\begin{tcolorbox}[title=Chinese \textsc{mT5-BASE}, colback=gray!5, colframe=gray!30, fonttitle=\bfseries, sharp corners, enhanced, breakable]
\footnotesize
\textbf{Prompt:}

\begin{flushleft}
\ttfamily
也反映了国内垂直电商的困境 \textbf{\texttt{<extra\_id\_0>}}平台型电商,垂直电商的 \textbf{\texttt{<extra\_id\_1>}}了,很难形成核心壁垒。”他说到。途棋牌在外贸方面,广东全年进出口顺差为1.54万亿元,出口增速快于进口4.5个百分点;一般贸易 \textbf{\texttt{<extra\_id\_2>}}比重为49.0\%,比上年提高2.0个百分点。从区域看 \textbf{\texttt{<extra\_id\_3>}}一带一路”沿线国家进出口总额增长6.3\%。 途 \textbf{\texttt{<extra\_id\_4>}} 傲头傲脑
\end{flushleft}

\vspace{0.5em}
\textbf{Reference:}
\begin{flushleft}
\ttfamily
\textbf{\texttt{<extra\_id\_0>}}。“相对 \textbf{\texttt{<extra\_id\_1>}}获客成本太高 \textbf{\texttt{<extra\_id\_2>}}占进出口总额的 \textbf{\texttt{<extra\_id\_3>}},对“ \textbf{\texttt{<extra\_id\_4>}}棋牌
\end{flushleft}

\vspace{0.5em}
\textbf{Prediction:}
\begin{flushleft}
\ttfamily
\textbf{\texttt{<extra\_id\_0>}}。“相对 \textbf{\texttt{<extra\_id\_1>}}获客成本太高 \textbf{\texttt{<extra\_id\_2>}}占进出口总额的 \textbf{\texttt{<extra\_id\_3>}},对“ \textbf{\texttt{<extra\_id\_4>}}棋牌
\end{flushleft}
\end{tcolorbox}
\end{CJK}

\begin{CJK}{UTF8}{min}
\begin{tcolorbox}[title=Japanese \textsc{mGPT-101}, colback=gray!5, colframe=gray!30, fonttitle=\bfseries, sharp corners, enhanced, breakable]
\footnotesize
\textbf{Prompt:}

\begin{flushleft}
\ttfamily
最高です。 義実家の姑・義姉は良い人なのですが、クーポンの服には出費を惜しまないためおすすめしていないと大変です。自分が惚れ込んだ物は用品が合わなくたって「いつか着れる」と買ってしまうので、用品がドンピシャの頃には収納に埋もれていたり、出してもアウトドアテーブル 120 80だって着たがらないんですよね。オーセンティックな感じの商品の服だと品質さえ良ければクーポンのことは考えなくて済むのに、カードの趣味
\end{flushleft}

\vspace{0.5em}
\textbf{Reference:}
\begin{flushleft}
\ttfamily
や私の反対意見などには耳も貸さずに購入するため、
\end{flushleft}

\vspace{0.5em}
\textbf{Prediction:}
\begin{flushleft}
\ttfamily
や私の反対意見などには耳も貸さずに購入するため、
\end{flushleft}
\end{tcolorbox}
\end{CJK}

\begin{CJK}{UTF8}{min}
\begin{tcolorbox}[title=Japanese \textsc{mT5-BASE}, colback=gray!5, colframe=gray!30, fonttitle=\bfseries, sharp corners, enhanced, breakable]
\footnotesize
\textbf{Promp t}

\begin{flushleft}
\ttfamily
の無料を聞いていない \textbf{\texttt{<extra\_id\_0>}}。用品が話しているときは夢中になるくせに、用品が念を押したことや予約 \textbf{\texttt{<extra\_id\_1>}}てしまうようです。アウトドアテーブル 120 80だって仕事だってひと通りこなしてきて、クーポンがないわけではないのですが、ポイントもない様子で、パソコンがいまいち噛み合わないのです。クーポンが \textbf{\texttt{<extra\_id\_2>}}言いませんが、サービスの妻はその傾向が強いです。 夏日になる日も増えてきましたが、私は昔からモバイル \textbf{\texttt{<extra\_id\_3>}}ダメで \textbf{\texttt{<extra\_id\_4>}}。この用品
\end{flushleft}

\vspace{0.5em}
\textbf{Reference:}
\begin{flushleft}
\ttfamily
\textbf{\texttt{<extra\_id\_0>}}と感じることが多いです \textbf{\texttt{<extra\_id\_1>}}はなぜか記憶から落ち \textbf{\texttt{<extra\_id\_2>}}みんなそうだとは \textbf{\texttt{<extra\_id\_3>}}が \textbf{\texttt{<extra\_id\_4>}}湿疹が出てしまいます
\end{flushleft}

\vspace{0.5em}
\textbf{Prediction:}
\begin{flushleft}
\ttfamily
\textbf{\texttt{<extra\_id\_0>}}と感じることが多いです \textbf{\texttt{<extra\_id\_1>}}はなぜか記憶から落ち \textbf{\texttt{<extra\_id\_2>}}みんなそうだとは \textbf{\texttt{<extra\_id\_3>}}が \textbf{\texttt{<extra\_id\_4>}}湿疹が出てしまいます
\end{flushleft}
\end{tcolorbox}
\end{CJK}

\clearpage
\subsection{Example of Unstable generation~\label{app:unstable-generation}}
\begin{tcolorbox}[
  title=Unstable Generation Example (\textsc{mT5-large}),
  colback=gray!5,
  colframe=gray!30,
  fonttitle=\bfseries,
  sharp corners,
  enhanced,
  breakable,
  width=\columnwidth
]
\footnotesize
\textbf{Reference:}

\ttfamily
\textbf{\texttt{<extra\_id\_0>}} beneficiaries of \textbf{\texttt{<extra\_id\_1>}} the \textbf{\texttt{<extra\_id\_2>}} the bond \textbf{\texttt{<extra\_id\_3>}}, agreeing to invest \textbf{\texttt{<extra\_id\_4>}} \$56.6 million in

\vspace{0.5em}
\textbf{Predicted:}

\textbf{\texttt{<extra\_id\_0>}} .   public bond..  mill  for parents school students vote  mill projects
\end{tcolorbox}

\subsection{Corpus Distribution}\label{app:corpus_distribution}
\noindent
\begin{minipage}{\textwidth}
  \centering
  \includegraphics[width=\textwidth]{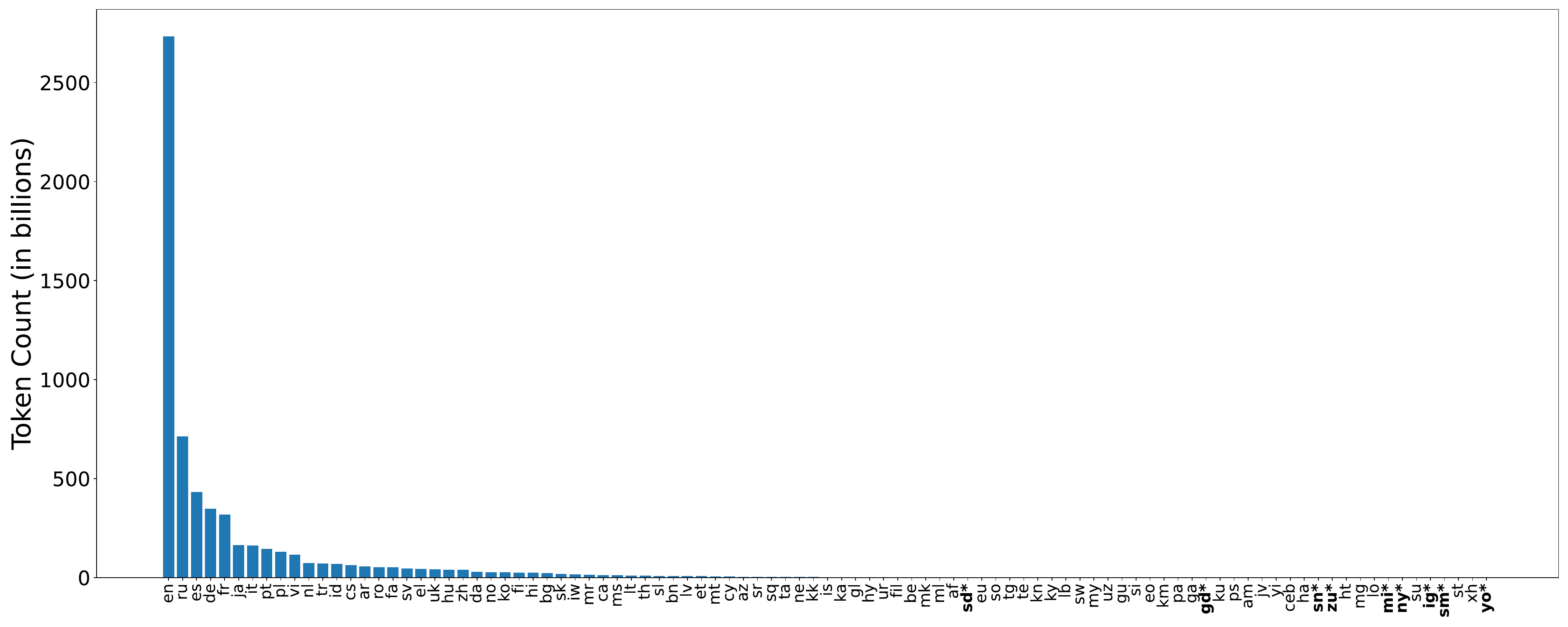}
  \captionof{figure}{\textsc{mGPT-101} \& \textsc{mT5} family analyzed language tokens distribution. The Languages marked with \textbf{*} have fewer than 50,000 sampled examples, averaging 33,960 examples per language.}
  \label{fig:corpus_distribution_1}
\end{minipage}

\vspace{1em}

\noindent
\begin{minipage}{\textwidth}
  \centering
  \includegraphics[width=\textwidth]{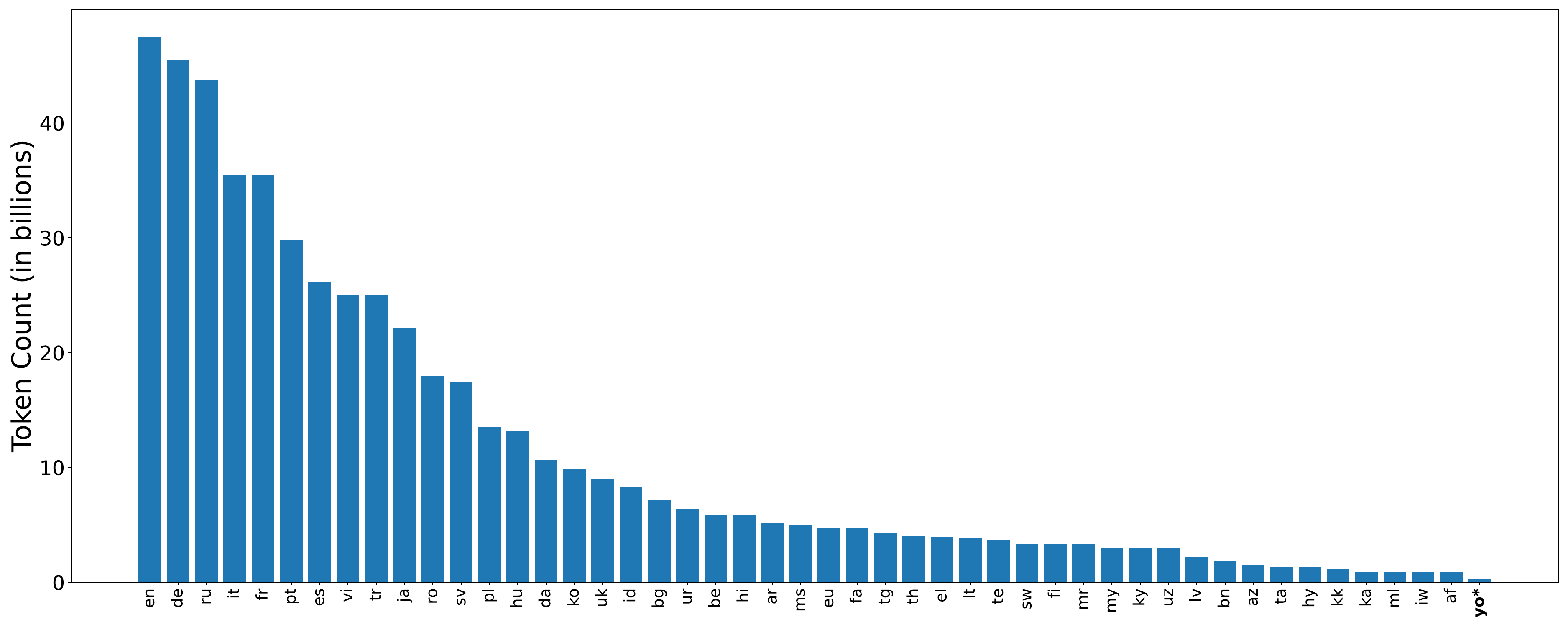}
  \captionof{figure}{\textsc{mGPT-61} (1.3B \& 13 B) family analyzed language tokens distribution. The language marked with \textbf{*} has fewer than 50,000 sampled examples, with a total of 17,339 examples.}
  \label{fig:corpus_distribution_2}
\end{minipage}

\end{document}